\pdfoutput=1
\documentclass{article}

\usepackage[preprint]{neurips_2026}

\usepackage[utf8]{inputenc} % allow utf-8 input
\usepackage[T1]{fontenc}    % use 8-bit T1 fonts
\usepackage{hyperref}       % hyperlinks
\usepackage{url}            % simple URL typesetting
\usepackage{booktabs}       % professional-quality tables
\usepackage{multirow}       % for multirow tables
\usepackage{amsmath}        % math environments and \text
\usepackage{amsfonts}       % blackboard math symbols
\usepackage{amssymb}        % \checkmark
\usepackage{dsfont}         % \mathds{1} indicator function
\usepackage{nicefrac}       % compact symbols for 1/2, etc.
\usepackage{microtype}      % microtypography
\usepackage[table]{xcolor}  % colors + \rowcolor
\usepackage{enumitem}       % customizable lists
\usepackage{graphicx}       % figures
\usepackage{pifont}         % dingbat check/cross symbols
\usepackage{caption}        % \captionof in minipage
\usepackage{algorithm}      % algorithm environment
\usepackage{algpseudocode}  % algorithmic pseudocode
\usepackage{wrapfig}        % text-wrapped tables/figures
\usepackage[most]{tcolorbox} % colored prompt boxes
\usepackage{titlesec}      % section/subsection spacing control (used in appendix)
\usepackage{stfloats}
\fnbelowfloat

\makeatletter
\renewcommand{\@noticestring}{}
\makeatother

\newcommand{\styledname}[1]{\texorpdfstring{\textsc{#1}}{#1}}
\newcommand{\methodname}{\styledname{ScaleCUA}}
\newcommand{\papertitle}{\methodname: Scaling Computer Use Agents with Verifiable Task Synthesis and Efficient Online RL}
\newcommand{\datapipelinename}{\styledname{VeriGen}}

\title{\papertitle}

\author{%
  Bowen Lv$^{1*\dagger}$, Xiao Liu$^{1,2*}$, Yanyu Ren$^{1\dagger}$, Hanyu Lai$^{1\dagger}$, Bohao Jing$^{2\dagger}$, \\
  \bfseries
  Hanchen Zhang$^{1\dagger}$, Yanxiao Zhao$^{1\dagger}$, Shuntian Yao$^{1\dagger}$, Jie Tang$^{1}$, Yuxiao Dong$^{1}$ \\[6pt]
  \normalfont
  \textsuperscript{1}Tsinghua University \qquad \textsuperscript{2}Z.AI
}

\begin{document}

\maketitle
\renewcommand{\thefootnote}{\fnsymbol{footnote}}
\footnotetext[1]{Equal contribution.}
\footnotetext[2]{Work done while these authors interned at Z.AI.}
\renewcommand{\thefootnote}{\arabic{footnote}}
\setcounter{footnote}{0}

% Page 1 is full; add a little height so the title footnotes fit at the bottom
% of page 1 rather than spilling to page 2.
\enlargethispage{2\baselineskip}

% Abstract
\begin{abstract}
Computer use agents (CUAs) are emerging as a powerful interface for automating complex digital workflows through visual perception and GUI execution. Online reinforcement learning with verifiable rewards (RLVR) has emerged as a key direction for scaling their capabilities. 
However, this paradigm is bottlenecked by verifiable data scarcity and online RL inefficiency.
To break these barriers, we introduce \textbf{\methodname{}}, a unified framework that scales online RL for CUAs via verifiable task synthesis and efficient training.
At the data level, we design \textbf{\datapipelinename{}}, an end-to-end framework for generating verifiable RL tasks through iterative docker interactions and a multi-agent feedback loop.
Scaled to 100+ concurrent agent workers via a shared docker interaction probe, this pipeline produces \textbf{24K+} verifiable tasks and nearly \textbf{3K} high-quality RL tasks.
To maximize sample efficiency, we propose Frontier Sampling, which tracks per-task capability and allocates rollouts to the current learning frontier.
On the training side, we further design Visual Context Segmentation, a sliding window over recent visual context that balances rollout and training-engine pressure, yielding a \textbf{2.83$\times$} training speedup over step-wise decomposition.
Together, \methodname{} achieves \textbf{68.7\%} on OSWorld and \textbf{54.0\%} on ScienceBoard, establishing new state-of-the-art performance among open-source computer use agents.
Code, models, and datasets are available at \url{https://github.com/THUDM/SCALE-CUA}.

\end{abstract}
\begin{figure}[!ht]
    \centering
    \begin{minipage}[t]{0.48\linewidth}
      \centering
      \includegraphics[height=0.62\linewidth,keepaspectratio]{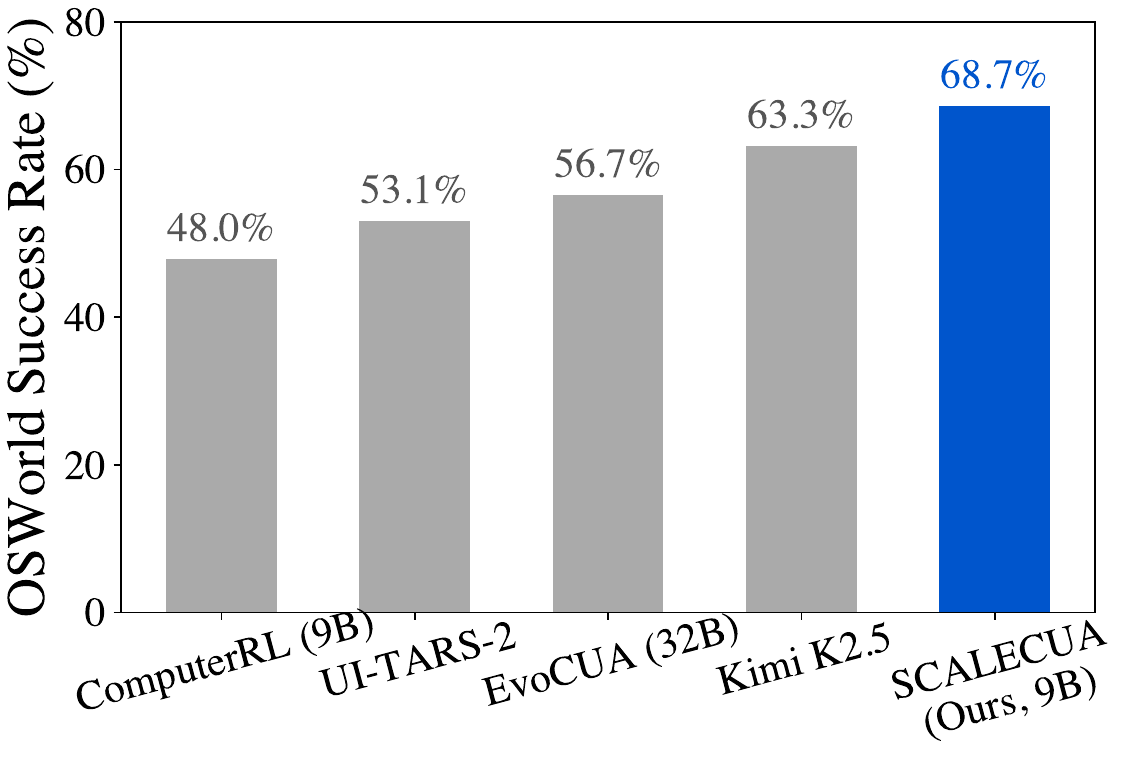}
    \end{minipage}
    \hfill
    \begin{minipage}[t]{0.48\linewidth}
      \centering
      \includegraphics[height=0.62\linewidth,keepaspectratio]{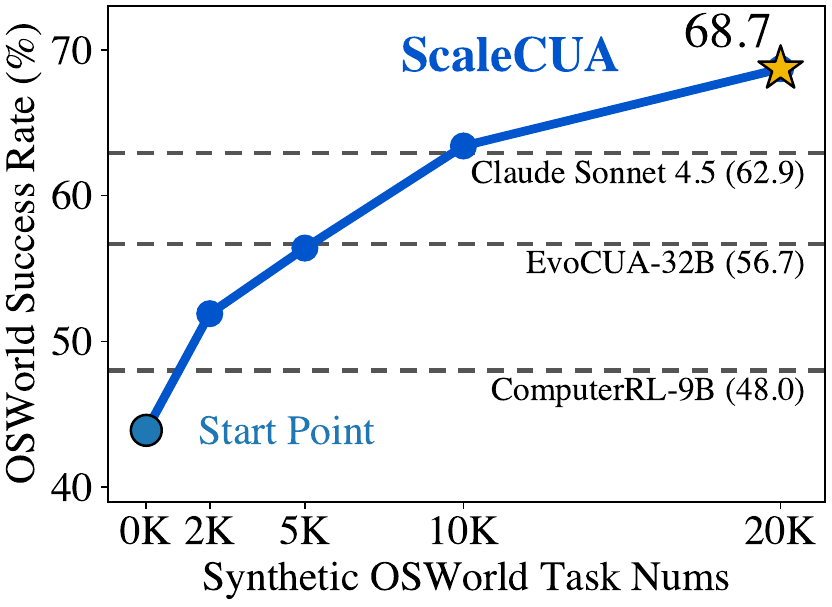}
    \end{minipage}
    \caption{\textbf{Left:} Comparison with open-source models on OSWorld. \methodname{} achieves 68.7\%, surpassing most prior open-source models including those 4$\times$ larger. \textbf{Right:} Effect of scaling generated verifiable tasks on OSWorld performance under the same training pipeline.}
    \label{fig:hero}
  \end{figure}

% Section 1: Introduction
\section{Introduction}
\label{sec:introduction}

\begin{figure}[t]
  \centering
  \setlength{\fboxsep}{0pt}\setlength{\fboxrule}{0.4pt}
  \newcommand{\tstep}[2]{{\scriptsize\textbf{Step~#1}\,$\cdot$\,#2}}
  \begin{minipage}[t]{0.323\textwidth}\centering
    \fbox{\includegraphics[width=\linewidth]{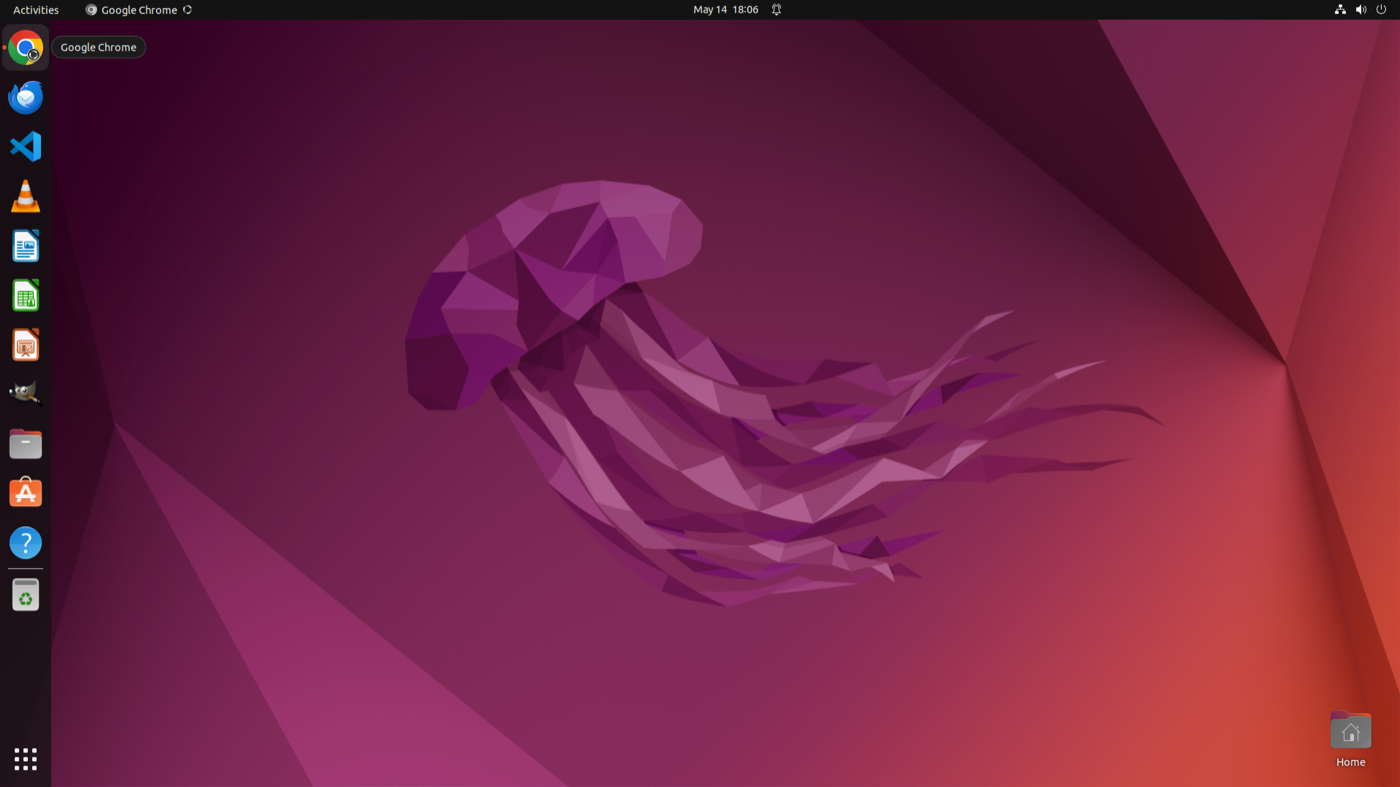}}\\[1.5pt]
    \tstep{1}{Read desktop state}
  \end{minipage}\hfill
  \begin{minipage}[t]{0.323\textwidth}\centering
    \fbox{\includegraphics[width=\linewidth]{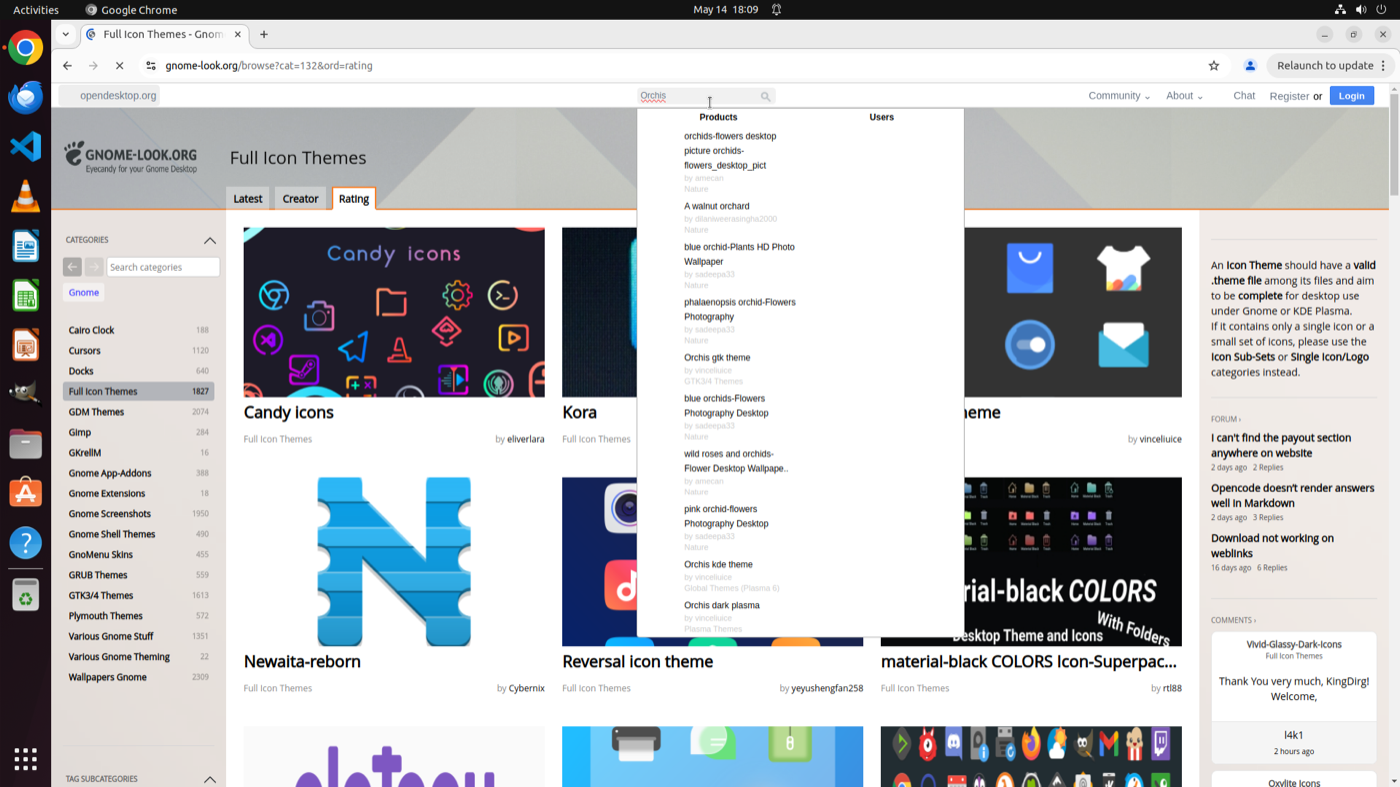}}\\[1.5pt]
    \tstep{8}{Browse gnome-look.org}
  \end{minipage}\hfill
  \begin{minipage}[t]{0.323\textwidth}\centering
    \fbox{\includegraphics[width=\linewidth]{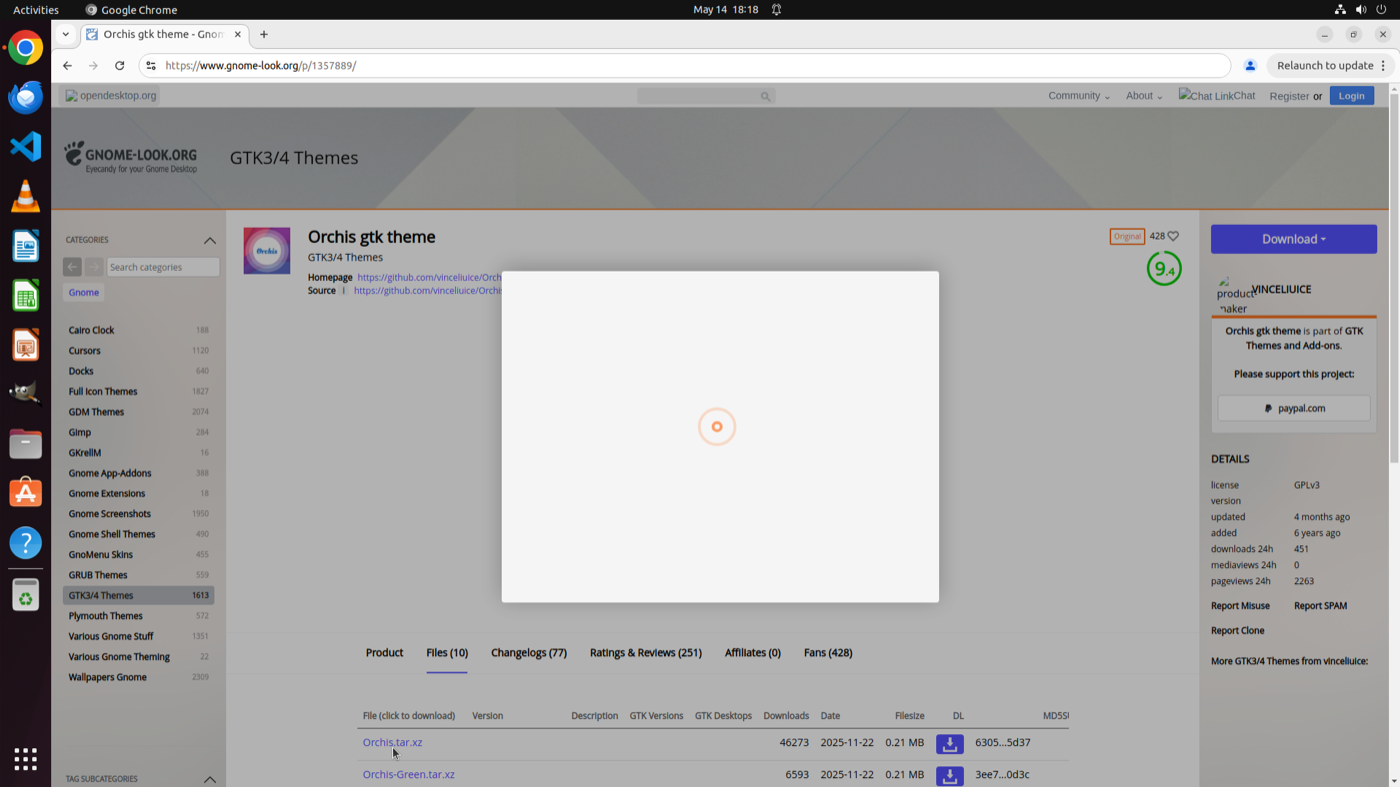}}\\[1.5pt]
    \tstep{27}{Download Orchis}
  \end{minipage}

  \vspace{4pt}
  \begin{minipage}[t]{0.323\textwidth}\centering
    \fbox{\includegraphics[width=\linewidth]{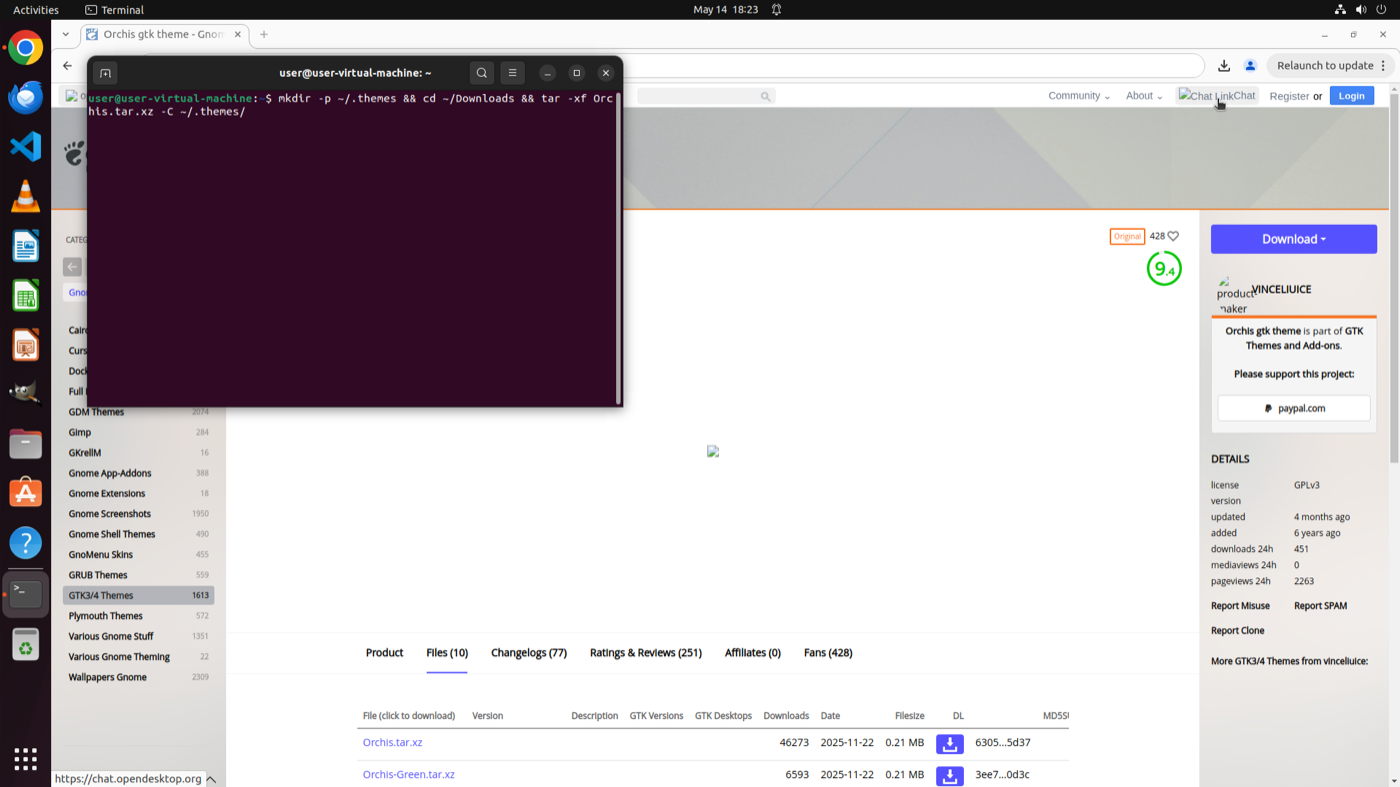}}\\[1.5pt]
    \tstep{34}{Extract in terminal}
  \end{minipage}\hfill
  \begin{minipage}[t]{0.323\textwidth}\centering
    \fbox{\includegraphics[width=\linewidth]{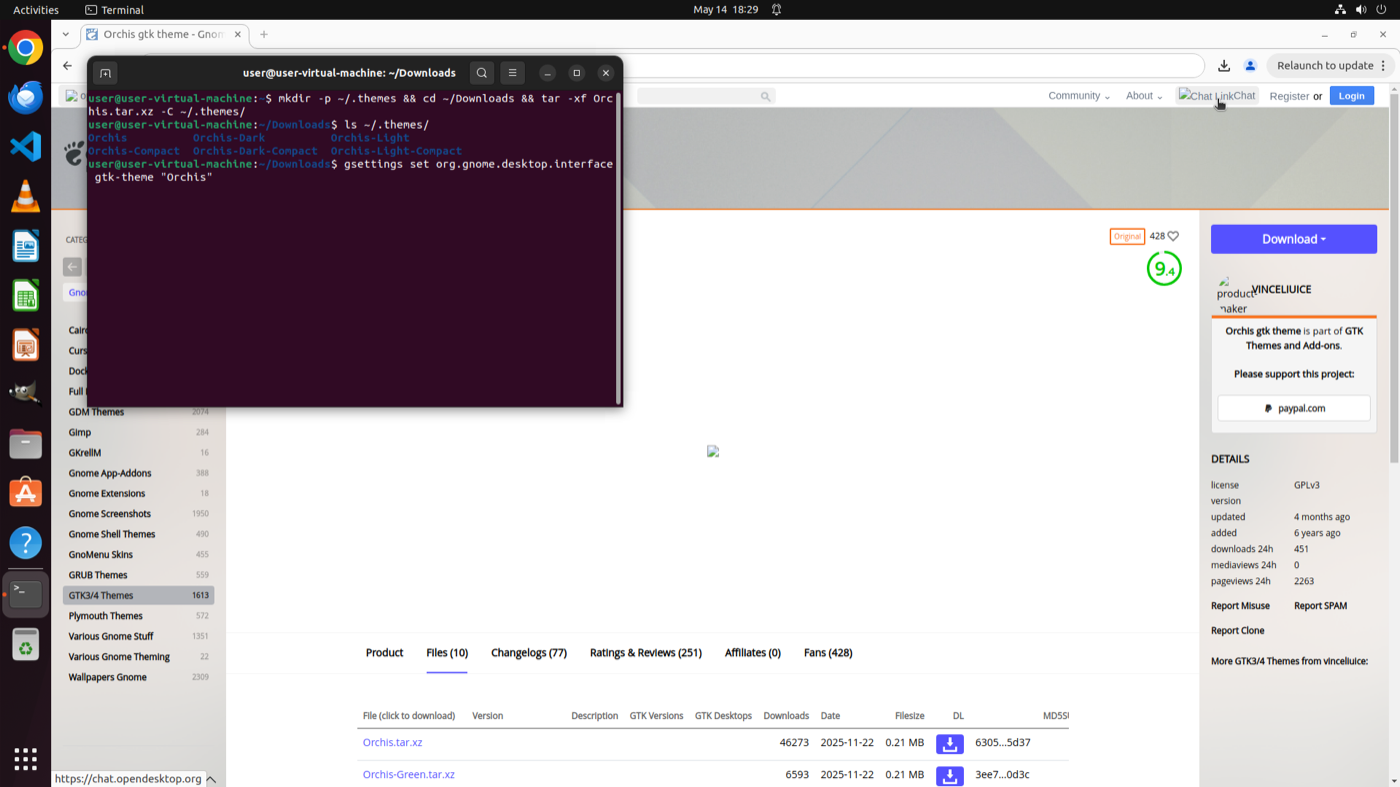}}\\[1.5pt]
    \tstep{38}{Apply via \texttt{gsettings}}
  \end{minipage}\hfill
  \begin{minipage}[t]{0.323\textwidth}\centering
    \fbox{\includegraphics[width=\linewidth]{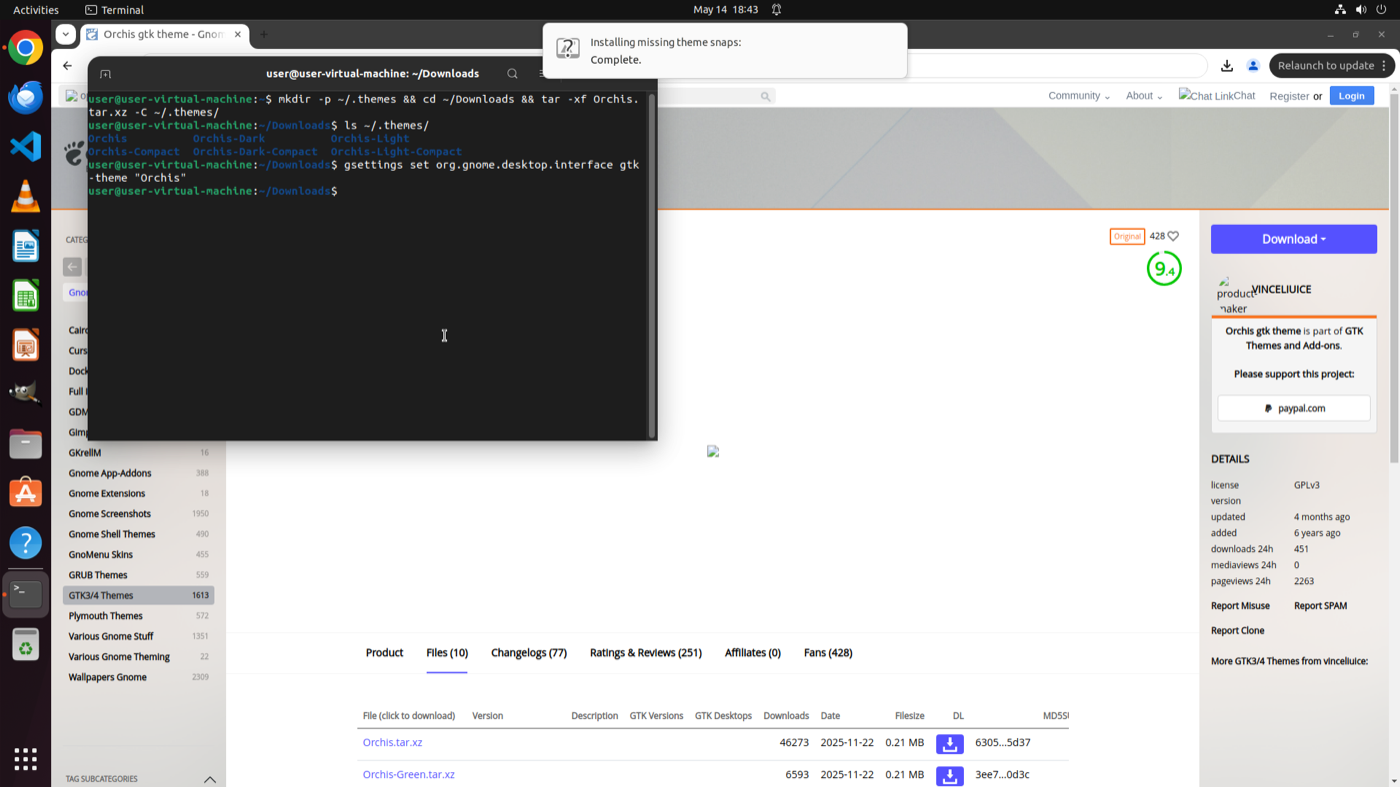}}\\[1.5pt]
    \tstep{47}{Theme applied, judge\,=\,1.0}
  \end{minipage}
  \caption{\textbf{Trained \methodname{} agent solving a 47-step multi-app OSWorld task end-to-end.} Given the instruction \emph{``install the Orchis GTK theme and switch to it for GNOME''}, the Qwen3.5-9B rollout progresses left-to-right, top-to-bottom---browsing gnome-look.org, downloading and extracting the theme, then applying it through the terminal and \texttt{gsettings}---until the OSWorld judge returns $\text{score}{=}1.0$.}
  \label{fig:teaser}
\end{figure}

Computer use agents (CUAs) have emerged as a transformative interface for automating everyday digital workflows through visual perception and GUI action execution~\citep{xie2024osworld, zhou2023webarena, hong2024cogagent, koh2024visualagentbench}. Unlike traditional automation built on structured APIs, CUAs operate directly on screen pixels and UI elements, enabling profound generalization across diverse applications and operating systems~\citep{gou2024uground, wu2024osatlas, qin2025uitars}. Each CUA task is naturally a multi-turn interaction loop: the agent observes a screenshot, acts on the interface, and iterates until task completion. Recent commercial breakthroughs such as Claude Computer Use~\citep{anthropic2024computeruse} and OpenAI Operator~\citep{openai2025operator} demonstrate the immense practical value of this paradigm. To further push the boundaries of CUA capabilities, the research community has begun shifting toward online reinforcement learning with verifiable rewards (RLVR)~\citep{bai2024digirl, qi2024webrl, lai2025computerrl, mobilerl2025, dart2025}.

However, while RLVR has shown initial promise, scaling it for CUAs remains largely unsolved due to verifiable data scarcity and online RL inefficiency.
On the data side, the bottleneck lies in producing verifiable and suitable GUI tasks at scale.
A verifiable GUI task is one paired with a deterministic judge function (e.g., a filesystem check, a browser-state inspection, or a piece of Python code) that scores the final environment state without any human or LLM judgment.
Unlike math~\citep{shao2024deepseekmath, guo2025deepseekr1, qwen25math2025} or code~\citep{jimenez2023swe}, where verifiers can be obtained easily, a GUI task naturally consists of only a live environment and an initial instruction, without a built-in mechanism to score the resulting state. This makes scaling verifiable GUI tasks fundamentally challenging.
A suitable GUI task is one that is neither trivially easy nor out-of-reach for the current model.
Under GRPO-style training, tasks where the model always succeeds or always fails contribute zero within-group advantage and waste the entire rollout group, so randomly synthesized tasks easily produce wasted rollouts at scale.
Prior work on automated GUI task synthesis has not addressed both axes---most efforts focus on trajectory-level data that is neither verifiable nor difficulty-controlled.

On the online RL side, inefficiency emerges in rollout sampling and multi-turn training. In terms of sampling, uniform task selection under GRPO-style training~\citep{shao2024deepseekmath} fails at both the per-batch and per-task level. At the per-batch level, uniform sampling frequently produces rollout groups in which all tasks either succeed or fail, which provides no useful signal for the RL training~\citep{yu2025dapo}. At the per-task level, the model cannot quickly handle specific tasks, as the sampling frequency of these tasks decreases when the task pool scales up.
In terms of training, the multi-turn GUI setting faces a rollout-training trade-off in how each rollout trajectory is packaged into training items. Trajectory-level training keeps the entire trajectory as a single training item, which accumulates all interaction screenshots but severely slows the rollout engine; the resulting long sequences also stress the training engine. Step-wise training instead emits one training item per interaction turn, each retaining only the final screenshot. This relieves the rollout engine, but the number of training items grows linearly with the average number of interaction turns, substantially slowing overall training.

To systematically break these barriers, we present \textbf{\methodname{}}, a unified framework for scaling computer use agents that combines verifiable task synthesis with efficient online RL.
For verifiable task synthesis, we design \textbf{\datapipelinename{}}, an end-to-end pipeline that takes both environment knowledge documents and model rollout trajectories as input, and iteratively interacts with live docker environments through a multi-agent feedback loop to refine each task and judge. For efficient sampling, we propose \textbf{Frontier Sampling}, which continuously tracks the sampling weight of each task and dynamically selects a batch of tasks best matched to the model's current capability before each rollout.
For multi-turn training, we propose \textbf{Visual Context Segmentation}, which keeps the recent visual context within a sliding window while preserving textual continuity. This controls visual-token growth during long-horizon training and avoids the sample-count inflation of step-wise decomposition, enabling more efficient optimization over large-scale multi-turn rollouts.

We apply \methodname{} to several leading vision-language models, including GLM-4.6V-Flash~\citep{glm45v2025}, Qwen3-VL-8B-Thinking~\citep{qwen3vl2025}, and Qwen3.5-9B~\citep{qwen32025}, observing consistent gains across all three models. With Qwen3.5-9B, \methodname{} achieves \textbf{68.7\%} on OSWorld~\citep{xie2024osworld} and \textbf{54.0\%} on ScienceBoard~\citep{sun2025scienceboard}, establishing new state-of-the-art results among open-source computer use agents. Our 9B model surpasses prior open-source baselines on OSWorld, including ComputerRL-9B (48.0\%), EvoCUA-32B (56.7\%), and Kimi K2.5 (63.3\%).

\begin{table}[!t]
\caption{Comparison of representative GUI-agent methods along the scaling dimensions.}
\label{tab:scaling_comparison}
\centering
\small
\newcommand{\cmark}{\textcolor{green!60!black}{\ding{51}}}
\newcommand{\xmark}{\textcolor{red!70!black}{\ding{55}}}
\setlength{\tabcolsep}{5pt}
\renewcommand{\arraystretch}{1.2}
\begin{tabular}{l c c c c}
\toprule
\textbf{Method}
  & \shortstack{\textbf{Verifiable}\\\textbf{Task Scaling}}
  & \shortstack{\textbf{Efficient}\\\textbf{Sampling}}
  & \shortstack{\textbf{Large-scale}\\\textbf{Online RL}}
  & \shortstack{\textbf{Efficient}\\\textbf{Multi-turn Training}} \\
\midrule
GUI-Genesis~\citep{guigenesis2026} & \cmark & \xmark & \xmark & \xmark \\
OS-Genesis~\citep{sun2025osgenesis} & \xmark & \xmark & \xmark & \xmark \\
OpenCUA~\citep{opencua2025} & \xmark & \xmark & \xmark & \xmark \\
ComputerRL~\citep{lai2025computerrl} & \xmark & \xmark & \cmark & \xmark \\
DART~\citep{dart2025} & \xmark & \cmark & \cmark & \xmark \\
MobileRL~\citep{mobilerl2025} & \xmark & \cmark & \cmark & \xmark \\
Mobile-Agent-v3.5~\citep{xu2026mobileagentv35} & \xmark & \cmark & \cmark & \xmark \\
ARPO~\citep{wang2025arpo} & \xmark & \xmark & \xmark & \xmark \\
EvoCUA~\citep{evocua2026} & \cmark & \xmark & \xmark & \xmark \\
\rowcolor{gray!15}
\methodname{} & \cmark & \cmark & \cmark & \cmark \\
\bottomrule
\end{tabular}
\end{table}

In summary, our main contributions are:
\begin{itemize}[leftmargin=*, itemsep=2pt, topsep=2pt]
    \item %\textbf{Scaling verifiable tasks via \datapipelinename{}.}
    We present \datapipelinename{}, an end-to-end framework that generates verifiable and suitable GUI tasks through iterative docker interactions and a multi-agent feedback loop. Through a shared docker interaction probe, \datapipelinename{} scales to 100+ agent workers and 100+ docker environments concurrently, yielding \textbf{24K+} verifiable tasks and nearly \textbf{3K} high-quality RL tasks that establish the data foundation for \methodname{}.
    \item %\textbf{Efficient sampling via Frontier Sampling.}
    We introduce Frontier Sampling, a lightweight data sampler that maintains per-task sampling weights and selects training tasks aligned with the model's current capability before each rollout. With small per-step overhead, it provides efficient learning signal in every batch and speeds up the model's progression from easier to harder tasks.
    \item %\textbf{Efficient multi-turn training via Visual Context Segmentation.}
    We propose Visual Context Segmentation, a sliding-window strategy that bounds visual-token accumulation while preserving textual continuity, balancing the pressure between rollout and training engines. This achieves a \textbf{2.83$\times$} end-to-end training speedup over step-wise decomposition without sacrificing cross-step reasoning.
\end{itemize}

% Section 2: Related Work
\section{Related Work}
\label{sec:related_work}

\noindent\textbf{Computer Use Benchmarks.}
Computer use agent evaluation has evolved from web navigation benchmarks such as Mind2Web~\citep{deng2023mind2web} and WebArena~\citep{zhou2023webarena} to full desktop environments such as WindowsAgentArena~\citep{bonatti2024windowsagentarena}, OSWorld~\citep{xie2024osworld}, and ScienceBoard~\citep{sun2025scienceboard}. OSWorld evaluates open-ended desktop tasks in real operating systems, while ScienceBoard focuses on scientific workflows involving professional software and domain-specific operations.

\noindent\textbf{Data Generation for GUI Agents.}
Recent systems expand agent data through task or trajectory synthesis, including AgentTrek~\citep{xu2024agenttrek}, AgentGen~\citep{chen2024agentgen}, AndroidGen~\citep{lai2025androidgen}, OpenCUA~\citep{opencua2025}, OS-Genesis~\citep{sun2025osgenesis}, GUI-Genesis~\citep{guigenesis2026}, and AgentSynth~\citep{agentsynth2025}.
These works reduce manual data construction, but many focus on reasoning trajectories, task proposals, mobile/web environments, or reconstructed lightweight settings rather than executable judge synthesis in full desktop OS environments.
GUI-Genesis is closely related in its use of code-native verifiable rewards, yet it generates tasks in reconstructed lightweight web environments.
\datapipelinename{} addresses this gap with an end-to-end framework that synthesizes verifiable GUI tasks through iterative interaction with live OS containers, scaling task generation far beyond manually curated pools.

\noindent\textbf{Online RL for GUI Agents.}
Online RL for GUI agents was pioneered by DigiRL~\citep{bai2024digirl} in live Android environments, then extended to web tasks by WebRL~\citep{qi2024webrl} with curriculum-based task selection.
Subsequent work brought online RL to desktop and mobile platforms, including ARPO~\citep{wang2025arpo}, DART~\citep{dart2025}, ComputerRL~\citep{lai2025computerrl}, EvoCUA~\citep{evocua2026}, MobileRL~\citep{mobilerl2025}, and multi-platform systems~\citep{mobileguirl2025, mobileagentv32025, xu2026mobileagentv35}, demonstrating that executable feedback can improve computer use models across environments.
However, these methods remain limited in scaling. Uniform sampling and fixed curricula waste expensive rollouts on tasks that are already solved or out of reach, while step-wise decomposition multiplies training items with interaction length, slowing the training engine~\citep{lai2025computerrl, dart2025, mobilerl2025}.
\methodname{} instead addresses these inefficiencies through Frontier Sampling for upstream task allocation and Visual Context Segmentation for balanced multi-turn training.

% Section 3: Method
\section{The \methodname{} Framework}
\label{sec:method}

\subsection{Overview}
\label{sec:overview}

\methodname{} is a unified framework that combines verifiable task generation and efficient online RL to scale computer use agents.
On the data side, \datapipelinename{} (\S\ref{sec:autogen}) synthesizes verifiable GUI tasks at scale through iterative interaction with live OS environments.
On the RL side, Frontier Sampling (\S\ref{sec:adaptive_sampling}) lightly tracks each task's sampling weight to direct rollouts at the model's current frontier, while Visual Context Segmentation (\S\ref{sec:sliding_window}) balances rollout and training engine pressure through sliding-window trajectory processing.

Concretely, the \methodname{} pipeline runs in three stages.
First, we train the base model on the initial verifiable task pool generated by \datapipelinename{}.
Next, we collect the trained model's rollouts on a subset of these tasks, and \datapipelinename{} applies trajectory-guided synthesis to convert these rollouts into a refined RL-ready task pool.
Finally, we run online RL on this refined pool with Frontier Sampling and Visual Context Segmentation.

\begin{figure}[t]
  \centering
  \includegraphics[width=0.9\linewidth]{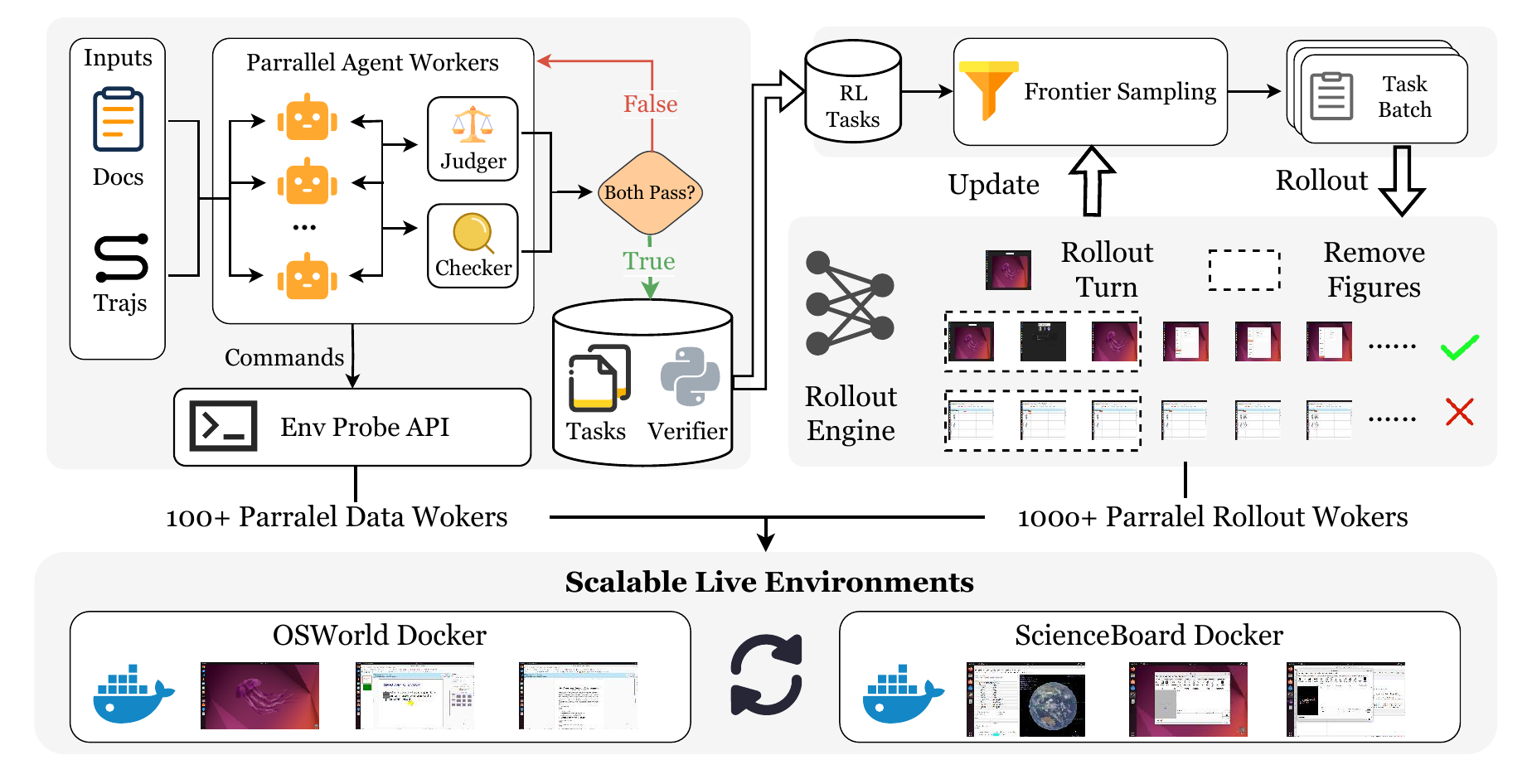}
  \caption{Overview of \methodname{}. \datapipelinename{} autonomously synthesizes verifiable GUI tasks with executable rewards from live OS environments at 100+ parallel worker scale; combined with upstream Frontier Sampling and long-horizon Visual Context Segmentation, \methodname{} enables efficient large-scale online RL for computer use agents.}
  \label{fig:overview}
\end{figure}

\subsection{\datapipelinename: Scaling Verifiable Task Pools}
\label{sec:autogen}

To solve the scarcity of verifiable GUI tasks, we build \datapipelinename{}, an end-to-end framework that synthesizes them by driving LLMs to interact with live OS containers and validating each judge through real execution. Appendix~\ref{app:verigen_task_example} shows an example of a generated task.

To produce verifiable GUI tasks, we build an iterative framework that combines environment interaction with a multi-agent feedback loop. Three independent LLM agents---proposer, judger, and checker---play distinct roles in this loop.
Starting from knowledge documents extracted per environment and a live docker container interface, the proposer agent generates an initial synthetic task with its verifier.
On one path, an independent judger agent performs a static semantic review to evaluate whether the task instruction is unambiguous and meaningful and whether the verifier's logic is theoretically sound, returning a verdict with reasoning.
On the other path, an independent checker agent runs a dynamic check by performing a ``dry run'' within the docker environment: it actively executes actions and observes the resulting state updates (screenshots and accessibility trees) to test whether the task goal is physically reachable and whether the verifier correctly evaluates the final outcome.
Combining both signals, the framework produces a final judgment on the candidate and decides whether to feed it back into the next loop.

To get more suitable tasks for the model's RL training, we further extend the framework with trajectory-guided task synthesis, taking the trained model's rollout trajectories and per-task pass rates as additional input.
For failed tasks, \datapipelinename{} inspects each step's action and the environment feedback to dynamically locate the earliest step at which the model starts to deviate from the original goal.
It then splits the original task into smaller subtasks, using each intermediate sub-goal as the target of a new, easier task with its own setup state and verifier.
For succeeded tasks, \datapipelinename{} clusters semantically similar tasks within each application (e.g., tasks operating on the same Excel sheet) and composes them into harder multi-objective tasks by chaining their goals and verifiers into a single instruction.

To support large-scale parallel data synthesis, we design a shared docker interaction probe that exposes an environment interface to agent workers.
Through this probe, agent workers retrieve screenshots and accessibility-tree information from the docker without entering it, while the probe executes dispatched actions inside the environment.
This separation lets \datapipelinename{} run 100+ agent workers concurrently against 100+ docker containers.
Combining iterative interaction, trajectory-guided synthesis, and the parallel probe, the full \datapipelinename{} framework produces \textbf{24K+} verifiable tasks and nearly \textbf{3K} high-quality RL tasks.

\subsection{Frontier Sampling over Large Task Pools}
\label{sec:adaptive_sampling}

Given the scaled task pool from \datapipelinename{}, online RL needs a more efficient sampling method for the model to converge faster.
We thus propose Frontier Sampling, a lightweight data sampler that maintains per-task sampling weights and dynamically allocates rollouts to tasks at the model's current capability frontier.

To track the model's capability on each task, we maintain an exponential moving average (EMA) of per-task success rates, which updates smoothly across iterations.
Before RL begins, we rollout the model on all tasks to obtain initial pass rates ($\hat{p}_i$) and filter out tasks outside $[p_{\min}, p_{\max}]$ (e.g., $[0.1, 0.9]$).
Each remaining task $i$ then has its EMA success rate $\hat{p}_i$ updated as:
\begin{equation}
    \hat{p}_i \leftarrow (1 - \alpha)\,\hat{p}_i + \alpha \cdot \tfrac{1}{n}\textstyle\sum_{j=1}^{n} \mathds{1}[\text{rollout } j \text{ succeeds}],
    \label{eq:ema_update}
\end{equation}
where $n$ rollouts are averaged after each RL iteration and $\alpha$ is the EMA smoothing factor (e.g., $\alpha{=}0.2$).
To convert these EMA rates into sampling probabilities in an appropriate way, we apply a Gaussian kernel distribution.
The Gaussian peaks at a target success rate, decays symmetrically on both sides, and drops off exponentially with distance, naturally directing rollouts to tasks at the model's current capability frontier.
Specifically, we compute the sampling weight $w_i$ as:
\begin{equation}
    w_i = \exp\!\Bigl(-\frac{(\hat{p}_i - \mu)^2}{2\sigma^2}\Bigr),
    \label{eq:sampling_weight}
\end{equation}

Here $\mu$ specifies the target success rate and $\sigma$ controls the bandwidth around it; we use $\mu{=}0.5$ and $\sigma{=}0.25$ in all experiments.
To preserve exploration, we additionally reserve a small fraction $\gamma$ ($\gamma$ = 0.2) of each batch for uniform sampling from the remaining tasks.
In practice, Frontier Sampling concentrates rollouts on tasks with the most useful learning signal, accelerating model convergence over uniform sampling and curriculum-style baselines.

\subsection{Visual Context Segmentation for Long-Horizon Multimodal RL}
\label{sec:sliding_window}
\begin{figure}[t]
  \centering
  \begin{minipage}[c]{0.58\linewidth}
    \scriptsize
    \hrule height 1pt
    \vspace{2pt}
    \noindent\textbf{Algorithm 1:} Sliding-Window Trajectory Segmentation with Token-ID Stream
    \vspace{2pt}
    \hrule height 0.4pt
    \vspace{2pt}
    \begin{algorithmic}[1]
    \Require Episode with $T$ turns; $K$ images to keep; $\Delta$ min removal threshold
    \Ensure Set of trace segments $\mathcal{S}$
    \State $\mathcal{S} \leftarrow \emptyset$;\; $n_{\text{img}} \leftarrow 0$;\; tokenize $\mathcal{H} \rightarrow (\mathbf{ids},\mathbf{mask}{=}\mathbf{0},\mathbf{logp}{=}\mathbf{0})$
    \For{$t = 1, \ldots, T$}
        \State Sample $(\mathbf{ids}^{\text{gen}}_t, \mathbf{logp}^{\text{gen}}_t) \sim \pi_\theta$;\; append with $\mathbf{mask}{=}\mathbf{1}$
        \State Execute action, receive $o_t$;\; $n_{\text{img}} \leftarrow n_{\text{img}} + 1$
        \If{$n_{\text{img}} > K + \Delta$}
            \State $\mathcal{S} \leftarrow \mathcal{S} \cup \{(\mathbf{ids},\mathbf{mask},\mathbf{logp})\}$;\; drop oldest $\Delta$ images from $\mathcal{H}$
            \State Re-tokenize $\mathcal{H} \rightarrow (\mathbf{ids},\mathbf{mask}{=}\mathbf{0},\mathbf{logp}{=}\mathbf{0})$
        \Else
            \State $\mathbf{diff} \leftarrow \textsc{Tokenize}(\mathcal{H} \oplus o_t)[|\textsc{Tokenize}(\mathcal{H})|{:}]$
            \State $\mathbf{ids} \mathrel{+}{=} \mathbf{diff}$;\; $\mathbf{mask} \mathrel{+}{=} \mathbf{0}$;\; $\mathbf{logp} \mathrel{+}{=} \mathbf{0}$
        \EndIf
    \EndFor
    \State $\mathcal{S} \leftarrow \mathcal{S} \cup \{(\mathbf{ids},\mathbf{mask},\mathbf{logp})\}$;\; assign reward $r$ to all segments
    \State \Return $\mathcal{S}$
    \end{algorithmic}
    \vspace{2pt}
    \hrule height 1pt
    \refstepcounter{algorithm}\label{alg:sliding_window_main}
  \end{minipage}\hfill
  \begin{minipage}[c]{0.40\linewidth}
    \centering
    \includegraphics[width=\linewidth]{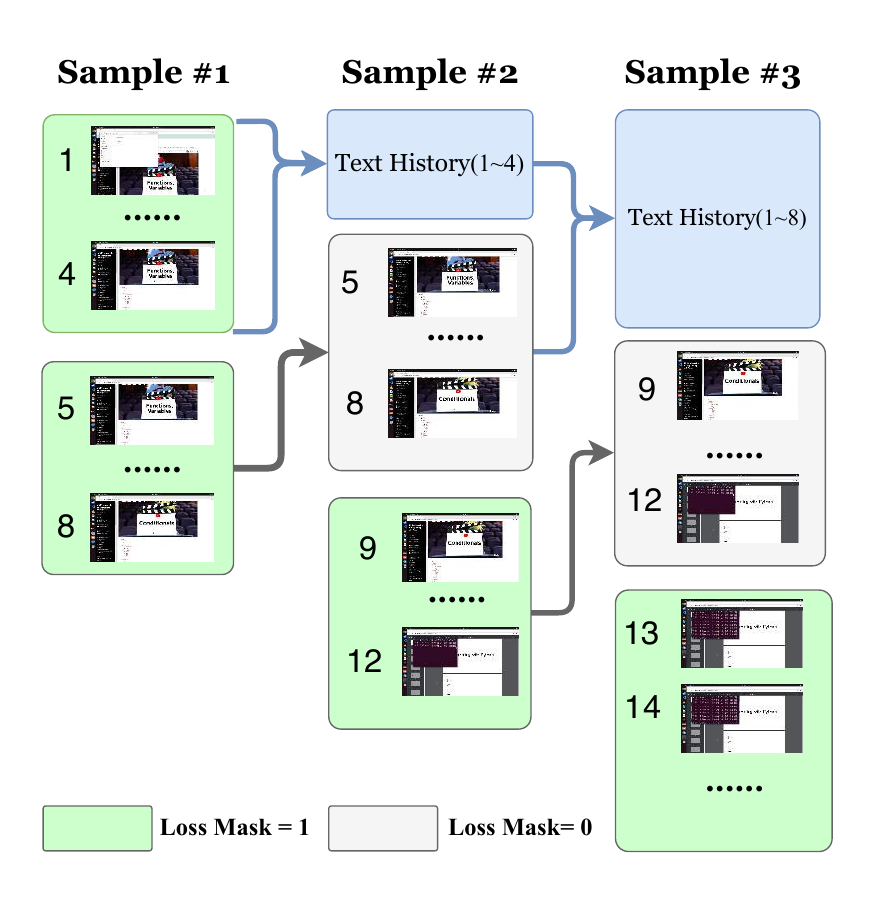}
  \end{minipage}
  \caption{Visual Context Segmentation. \textbf{Left:} sliding-window trajectory processing with a token-ID stream. \textbf{Right:} when the visual window overflows, the oldest screenshots are dropped and replaced by a text summary; trainable masks (green, $\text{loss}{=}1$) cover only assistant responses, while prior tokens stay as context (gray, $\text{loss}{=}0$).}
  \label{fig:sliding_window}
  \vspace{-2mm}
\end{figure}

Scaling online RL for CUAs critically depends on training efficiency.
In the multi-turn GUI setting, keeping the full visual sequence places a heavy burden on the rollout engine and slows inference dramatically, while splitting at every screenshot inflates training items and slows parameter updates.
\begin{wrapfigure}{r}{0.35\linewidth}
\vspace{-0.8em}
\centering
\includegraphics[width=\linewidth]{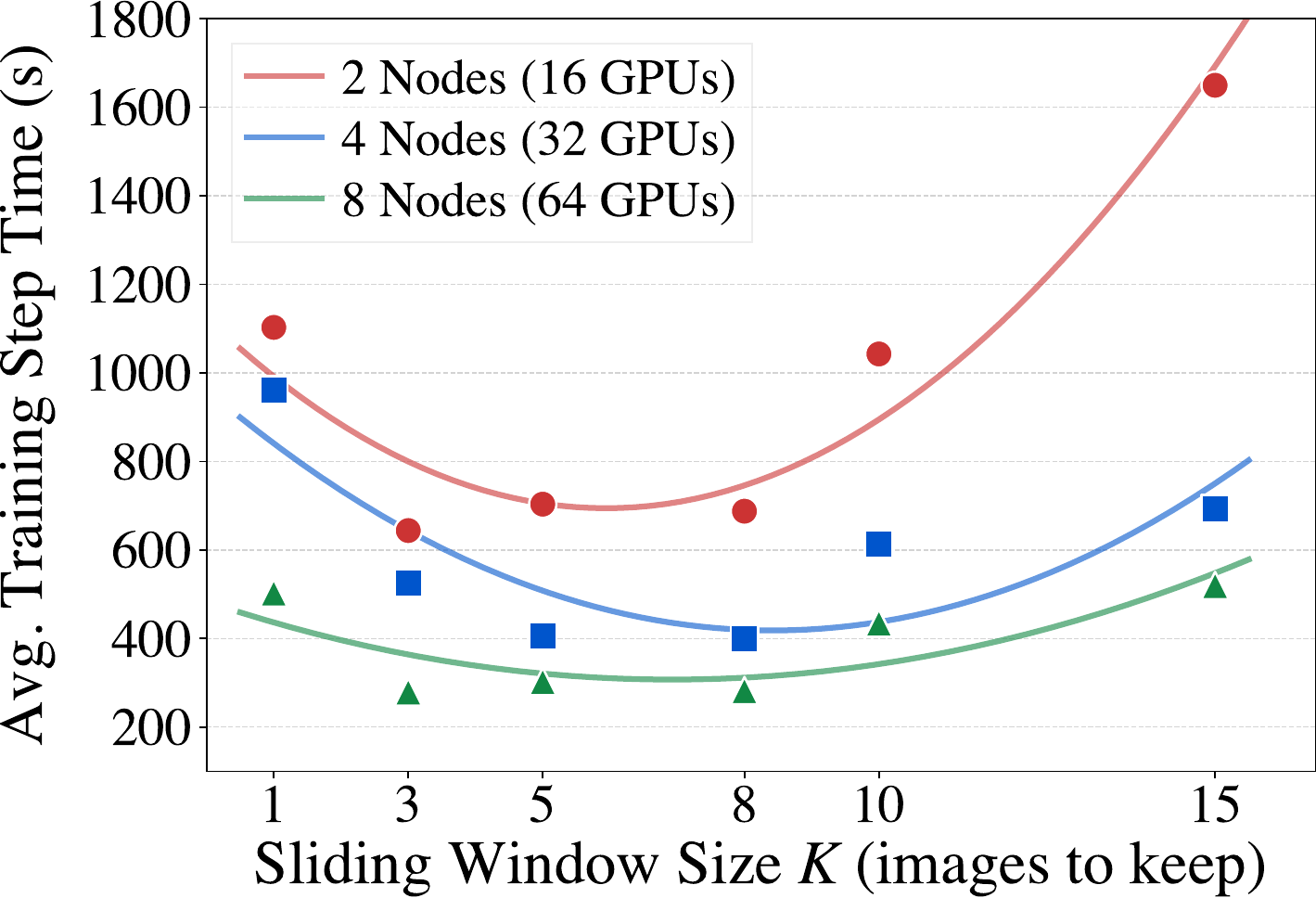}
\caption{Sliding-Window training step time across $K$ at 2/4/8-node scales; moderate $K$ is optimal.}
\label{fig:sliding_speedup}
\vspace{-0.8em}
\end{wrapfigure}
To balance this trade-off, we propose Visual Context Segmentation, which keeps the recent visual context within a sliding window while preserving textual continuity. This simultaneously bounds the per-rollout visual-token load on the rollout engine and the training-item count on the training engine, yielding an efficient multi-turn training scheme without sacrificing cross-step reasoning.

Consider a training batch of $n$ trajectories with an average length of $T$ interaction turns. We define $K$ as the number of retained screenshots (the window size) and $\Delta$ as the minimum removal threshold (normally $K{=}\Delta$). The naive trajectory-level mode produces $O(n)$ training items and step-wise decomposition produces $O(nT)$, while Visual Context Segmentation produces $O(nT/K)$.

In practice, we maintain a token-ID stream (the running concatenation of input IDs, loss masks, and rollout log-probabilities) and a text history stream that preserves the full conversation with images during the rollout process.
At each turn, if the text history holds fewer than $K{+}\Delta$ screenshots, the new observation and assistant response are appended to the token-ID stream via token-diff.
Once the count reaches $K{+}\Delta$, pruning triggers: the current stream is saved as a training item, the oldest $\Delta$ screenshots are removed, and the remaining history is re-tokenized into a fresh token-ID stream.
Figure~\ref{fig:sliding_window} summarizes the procedure; full details of the token-ID stream are provided in Appendix~\ref{app:sliding_algorithm}.

To pick $K$ and $\Delta$, we run a pre-experiment with all training hyperparameters fixed and sweep $K{=}\Delta\in\{1,3,5,8,10,15\}$ across 2/4/8-node compute scales (Figure~\ref{fig:sliding_speedup}).
Capping rollout length at 50 turns, $K{=}5$--$8$ reaches the local optimum at every scale, balancing per-segment visual-token pressure on the rollout engine against item-count overhead on the training engine.
Under this setting, sliding-window training delivers a \textbf{2.83$\times$} end-to-end speedup over step-wise decomposition.

% Section 4: Experiments
\section{Experiments}
\label{sec:experiments}
\subsection{Setup}
\label{sec:exp_setup}
We apply \methodname{} to three models---GLM-4.6V-Flash~\citep{glm45v2025}, Qwen3-VL-8B-Thinking~\citep{qwen3vl2025}, and Qwen3.5-9B~\citep{qwen32025}---and evaluate on OSWorld~\citep{xie2024osworld} and ScienceBoard~\citep{sun2025scienceboard}.
Our agent operates from screenshot observations and uses each benchmark's exposed computer-action interface: OSWorld uses mouse and keyboard actions, while ScienceBoard additionally provides task-completion, answer-submission, and file-writing actions (see Appendix~\ref{app:action_space} for the full action space).

We build the training framework on top of AgentRL~\citep{agentrl2025} and adapt it for multimodal CUA training, using vLLM~\citep{kwon2023efficient} as the rollout engine and Megatron-LM~\citep{shoeybi2019megatron} as the training engine.
Algorithmically, we use GRPO~\citep{shao2024deepseekmath} with DAPO-style asymmetric clipping~\citep{yu2025dapo} and a KL penalty against the reference policy, integrating Frontier Sampling for task selection and Visual Context Segmentation for trajectory processing (full objective in Appendix~\ref{app:training_details}).
Detailed hyperparameters and infrastructure configurations are provided in Appendix~\ref{app:training_details}.

\begin{table}[!htbp]
\caption{OSWorld success rates (\%). Best open-source single-model results are \textbf{bolded}; ours are shaded. Subscripts indicate the standard deviation across 4 independent rollout passes per task. Rows without per-subset scores are extracted directly from the corresponding technical reports.}
\label{tab:osworld}
\centering
\resizebox{\textwidth}{!}{%
\begin{tabular}{l r c c c c c c}
\toprule
\textbf{Model} & \textbf{Steps} & \textbf{OS} & \textbf{Office} & \textbf{Daily} & \textbf{Professional} & \textbf{Workflow} & \textbf{Overall} \\
\midrule
Human & -- & -- & -- & -- & -- & -- & 72.4 \\
\midrule
\multicolumn{8}{l}{\emph{Proprietary models}} \\
GPT-5.4~\citep{openai2025gpt}$^{\dagger\S}$                    & --  & -- & -- & -- & -- & -- & 75.0 \\
Claude Opus 4.6~\citep{anthropic2025claude}$^{\dagger\S}$       & --  & -- & -- & -- & -- & -- & 72.7 \\
Claude Sonnet 4.5~\citep{anthropic2025claude}$^\dagger$          & 100 & 70.8 & 72.6 & 61.4 & 63.2 & 49.5 & 62.9 \\
Claude Sonnet 4~\citep{anthropic2025claude}$^\dagger$            & 50  & 45.8 & 43.5 & 55.1 & 55.1 & 28.5 & 43.9 \\
Seed-1.8~\citep{seed2026seed1}$^\dagger$                           & 100 & 66.7 & 68.8 & 67.1 & 71.4 & 42.4 & 61.9 \\
OpenAI CUA~\citep{openai2025operator}$^\dagger$                  & 50  & 70.8 & 23.0 & 37.2 & 51.0 & 15.8 & 31.3 \\
CoACT-1~\citep{hou2024coact}$^\ddagger$                            & 50  & 70.8 & 60.6 & 54.1 & 69.4 & 42.4 & 56.4 \\
Agent S2.5 w/ o3~\citep{agents22025}$^\ddagger$                  & 50  & 75.0 & 53.8 & 54.0 & 75.5 & 39.5 & 54.2 \\
GTA1 w/ o3~\citep{gta12025}$^\ddagger$                           & 100 & 62.5 & 54.6 & 60.3 & 61.2 & 38.3 & 53.1 \\
Jedi-7B w/ o3~\citep{jedi2025}$^\ddagger$                        & 50  & 54.2 & 47.0 & 62.1 & 69.4 & 35.0 & 50.6 \\
\midrule
\multicolumn{8}{l}{\emph{Open-source models}} \\
Kimi K2.5~\citep{kimik252026}                                    & 100 & 73.9 & 69.2 & 66.4 & 73.5 & 46.1 & 63.3 \\
EvoCUA-32B~\citep{evocua2026}                                    & 50  & 78.3 & 59.8 & 64.5 & 81.6 & 27.9 & 56.7 \\
Mobile-Agent-v3.5~\citep{xu2026mobileagentv35}$^\S$             & --  & -- & -- & -- & -- & -- & 56.5 \\
UI-TARS-2~\citep{uitars22025}                                    & 100 & 41.7 & 61.1 & 62.1 & 61.2 & 34.1 & 53.1 \\
ComputerRL-9B~\citep{lai2025computerrl}                          & 50  & 75.0 & 45.8 & 52.5 & 71.4 & 27.9 & 48.0 \\
OpenCUA-72B~\citep{opencua2025}                                  & 50  & 45.8 & 42.7 & 53.8 & 75.5 & 22.3 & 44.9 \\
OpenCUA-32B~\citep{opencua2025}                                  & 50  & 60.9 & 29.9 & 41.0 & 65.3 & 14.6 & 35.1 \\
DART-GUI-7B~\citep{dart2025}                                     & 30  & 62.5 & 39.3 & 50.8 & 73.5 & 16.7 & 42.1 \\
Qwen3.5-9B~\citep{qwen32025}$^\S$                               & --  & -- & -- & -- & -- & -- & 41.8 \\
Qwen3-VL-8B-Thinking~\citep{qwen3vl2025}$^\S$                   & --  & -- & -- & -- & -- & -- & 33.9 \\
\midrule
\multicolumn{8}{l}{\emph{Ours}} \\
\rowcolor{gray!15}
\methodname{}-GLM-4.6V-Flash$^\dagger$ & 50 & \textbf{83.3}$_{\pm 4.0}$ & \textbf{71.8}$_{\pm 2.7}$ & 64.0$_{\pm 1.9}$ & 85.5$_{\pm 1.1}$ & 46.9$_{\pm 2.4}$ & 66.5$_{\pm 0.5}$ \\
\rowcolor{gray!15}
\methodname{}-Qwen3-VL-8B & 50 & \textbf{83.3}$_{\pm 7.1}$ & 68.4$_{\pm 1.4}$ & \textbf{79.4}$_{\pm 0.8}$ & 81.6$_{\pm 2.8}$ & 45.2$_{\pm 3.3}$ & 67.7$_{\pm 1.2}$ \\
\rowcolor{gray!15}
\methodname{}-Qwen3.5-9B & 50 & 79.2$_{\pm 7.2}$ & 70.6$_{\pm 2.2}$ & 74.3$_{\pm 3.2}$ & \textbf{89.5}$_{\pm 3.6}$ & \textbf{48.1}$_{\pm 1.6}$ & \textbf{68.7}$_{\pm 0.7}$ \\
\bottomrule
\end{tabular}%
}
\end{table}

\subsection{Main Results}

\label{sec:main_results}
\noindent\textbf{OSWorld.} Table~\ref{tab:osworld} reports per-domain results on OSWorld, where we compare \methodname{} against proprietary systems (GPT-5.4, Claude Opus 4.6 / Sonnet 4.5, Seed-1.8, OpenAI CUA) and open-source baselines (Kimi K2.5, EvoCUA-32B, ComputerRL-9B, UI-TARS-2).
\methodname{}-Qwen3.5-9B achieves \textbf{68.7\%}, surpassing the strongest open-source model Kimi K2.5 (63.3\%) and proprietary systems Claude Sonnet 4.5 (62.9\%) and Seed-1.8 (61.9\%), with gains most pronounced in the long-horizon Professional (\textbf{89.5\%}) and Workflow (\textbf{48.1\%}) categories.
All three models obtain consistent absolute gains over their base models (e.g., Qwen3.5-9B: 41.8\%${\to}$68.7\%, $+\textbf{26.9}$), confirming that \methodname{} is model-agnostic.

\begin{table}[!htbp]
\caption{ScienceBoard success rates (\%). Best open-source single-model results are \textbf{bolded}; ours are shaded. Subscripts indicate the standard deviation across 4 independent rollout passes per task.}
\label{tab:scienceboard}
\centering
\resizebox{\textwidth}{!}{%
\begin{tabular}{l l c c c c c c c}
\toprule
\textbf{Model} & \textbf{Obs.} & \textbf{Alg} & \textbf{Biochem} & \textbf{GIS} & \textbf{ATP} & \textbf{Astron} & \textbf{Doc} & \textbf{Overall} \\
\midrule
Human & -- & 74.2 & 69.0 & 55.9 & 42.3 & 51.5 & 68.8 & 60.3 \\
\midrule
\multicolumn{9}{l}{\emph{Proprietary models}} \\
Claude Opus 4.6~\citep{anthropic2025claude}$^\dagger$   & screen & 64.5 & 55.2 & 38.2 & 15.4 & 66.7 & 87.5 & 52.7 \\
Claude 3.7 Sonnet~\citep{anthropic2025claude}$^\dagger$  & s+a11y & 12.9 & 41.4 & 8.8 & 3.9 & 9.1 & 18.8 & 15.8 \\
GPT-4o~\citep{openai2025gpt}$^\dagger$                  & s+a11y & 22.6 & 37.9 & 2.9 & 7.7 & 3.0 & 12.5 & 14.4 \\
Gemini-2.0-Flash$^\dagger$                               & s+a11y & 16.1 & 24.1 & 2.9 & 0.0 & 18.2 & 12.5 & 12.3 \\
o3-mini~\citep{openai2025gpt}$^\dagger$                  & a11y   & 16.1 & 20.7 & 2.9 & 3.9 & 15.2 & 6.2 & 10.8 \\
\midrule
\multicolumn{9}{l}{\emph{Open-source models}} \\
GPT-4o $\rightarrow$ GUI-Actor$^\ddagger$                & screen & 21.9 & 44.8 & 2.9 & -- & 12.1 & -- & 20.4 \\
GPT-4o $\rightarrow$ Qwen2.5-VL-7B$^\ddagger$           & screen & 12.5 & 34.5 & 11.8 & -- & 9.1 & -- & 17.0 \\
Qwen2.5-VL-72B                                           & screen & 22.6 & 27.6 & 5.9 & 0.0 & 9.1 & 12.5 & 12.9 \\
Qwen3-VL-225B~\citep{qwen3vl2025}                       & screen & 6.5  & 17.2 & 2.9 & 0.0 & 9.1 & 0.0 & 6.5 \\
UI-TARS-1.5~\citep{qin2025uitars}                       & screen & 12.9 & 13.8 & 0.0 & 0.0 & 6.1 & 0.0 & 5.9 \\
InternVL3-78B                                             & s+a11y & 6.5  & 3.5  & 0.0 & 0.0 & 3.0  & 6.2 & 3.2  \\
\midrule
\multicolumn{9}{l}{\emph{Ours}} \\
\rowcolor{gray!15}
\methodname{}-GLM-4.6V-Flash$^\dagger$ & screen & 67.7$_{\pm 4.6}$ & 60.3$_{\pm 2.8}$ & \textbf{39.1}$_{\pm 6.5}$ & 8.0$_{\pm 4.0}$ & 60.6$_{\pm 3.9}$ & 57.8$_{\pm 5.3}$ & 49.4$_{\pm 2.8}$ \\
\rowcolor{gray!15}
\methodname{}-Qwen3-VL-8B & screen & \textbf{71.0}$_{\pm 6.7}$ & 58.6$_{\pm 2.8}$ & 38.2$_{\pm 2.8}$ & 15.0$_{\pm 3.1}$ & 59.4$_{\pm 3.9}$ & 62.5$_{\pm 13.3}$ & 51.9$_{\pm 1.5}$ \\
\rowcolor{gray!15}
\methodname{}-Qwen3.5-9B & screen & 67.7$_{\pm 5.6}$ & \textbf{62.1}$_{\pm 5.2}$ & 35.3$_{\pm 2.9}$ & \textbf{19.0}$_{\pm 1.7}$ & \textbf{60.6}$_{\pm 3.9}$ & \textbf{86.7}$_{\pm 8.2}$ & \textbf{54.0}$_{\pm 0.5}$ \\
\bottomrule
\end{tabular}%
}
\end{table}

\noindent\textbf{ScienceBoard.} To verify whether the pipeline remains effective on professional, knowledge-intensive software, we further evaluate on ScienceBoard~\citep{sun2025scienceboard}, which spans symbolic algebra (KAlgebra), molecular visualization (ChimeraX), geographic information systems (GrassGIS), Lean theorem proving, astronomical simulation (Celestia), and scientific document authoring (TeXstudio).
\methodname{}-Qwen3.5-9B reaches \textbf{54.0\%} (Table~\ref{tab:scienceboard}), surpassing Claude Opus 4.6 (52.7\%) and establishing a new state-of-the-art, with the largest gains on TeXstudio (\textbf{86.7\%}) and Lean theorem proving (\textbf{19.0\%}, vs.\ $\le 15.4\%$ for all proprietary baselines), confirming that \datapipelinename{} can synthesize verifiable tasks even for professional scientific software.

\subsection{Ablation Studies}
\label{sec:ablation}

% To isolate the contribution of each component, we conduct ablation studies on OSWorld using Qwen3.5-9B~\citep{qwen32025} as the base model. We perform detailed ablations on \datapipelinename{} and the two training-side mechanisms to demonstrate the importance of each.

To isolate each \methodname{} component, we report OSWorld scores when individual components are removed from the full pipeline (Table~\ref{tab:component_ablation}). Starting from full \methodname{} (\textbf{68.7\%}), removing \datapipelinename{} entirely drops the model to its 43.9\% base score, while removing Frontier Sampling or Visual Context Segmentation alone yields 63.7\% and 62.2\% respectively, showing that all three components contribute substantively and complementarily.

\begin{table}[!htbp]
\centering
\caption{\textbf{Left:} Component ablation of \methodname{} on OSWorld with Qwen3.5-9B backbone. \textbf{Right:} Executable rate (\%) of generated judge functions when each role is removed from the \datapipelinename{} multi-agent loop.}
\label{tab:ablation_combined}
\newcommand{\gain}[1]{\,{\scriptsize\textcolor{red!75!black}{#1}}}
\begin{minipage}[t]{0.62\linewidth}
\vspace{0pt}
\centering
\small
\begin{tabular}{@{}lc@{}}
\toprule
Configuration & OSWorld (\%) \\
\midrule
Full \methodname{}                        & \textbf{68.7} \\
\ \ w/o \datapipelinename{} (Base)         & 43.9 \\
\ \ w/o Frontier Sampling                  & 63.7 \\
\ \ w/o Visual Context Segmentation        & 62.2 \\
\bottomrule
\end{tabular}
\label{tab:component_ablation}
\end{minipage}\hfill
\begin{minipage}[t]{0.35\linewidth}
\vspace{0pt}
\centering
\small
\begin{tabular}{@{}lc@{}}
\toprule
Configuration & Exec. rate \\
\midrule
Full pipeline           & \textbf{94.5} \\
\ \ w/o LLM Judge Agent & 62.3 \\
\ \ w/o Fix Agent       & 78.1 \\
\ \ w/o Rule Validator  & 86.2 \\
\bottomrule
\end{tabular}
\label{tab:autogen_role_ablation}
\end{minipage}
\end{table}

\noindent\textbf{\datapipelinename{} role ablation.} We isolate the contribution of the multi-agent filtering roles by removing each module from the generation loop. As shown in Table~\ref{tab:autogen_role_ablation}, the full pipeline achieves a 94.5\% executable rate for generated judges. Removing the LLM Judge Agent causes the largest drop ($-$32.2\%), followed by the Fix Agent ($-$16.4\%) and Rule Validator ($-$8.3\%), indicating that independent checking and repair roles are important for reliable executable reward construction.
\begin{figure}[!t]
  \centering
  \begin{minipage}[t]{0.58\linewidth}
    \centering
    \includegraphics[width=\linewidth]{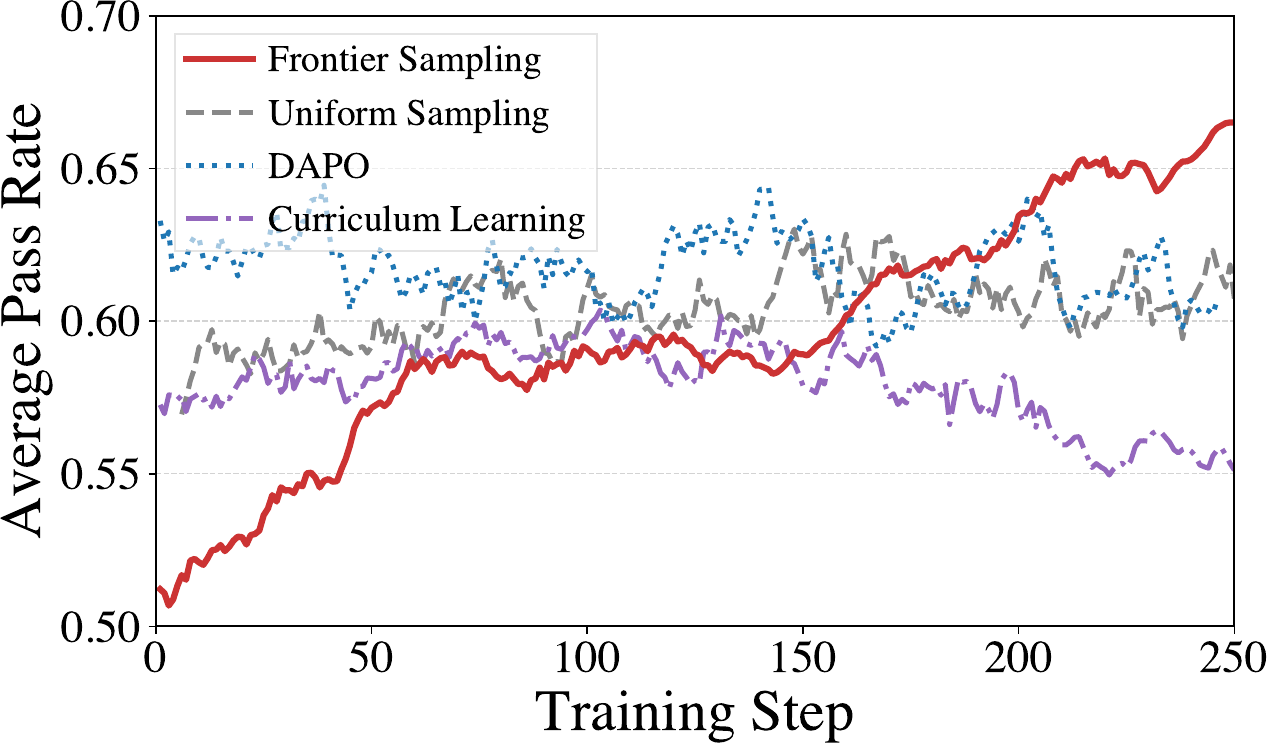}
  \end{minipage}\hfill
  \begin{minipage}[t]{0.40\linewidth}
    \centering
    \includegraphics[width=\linewidth]{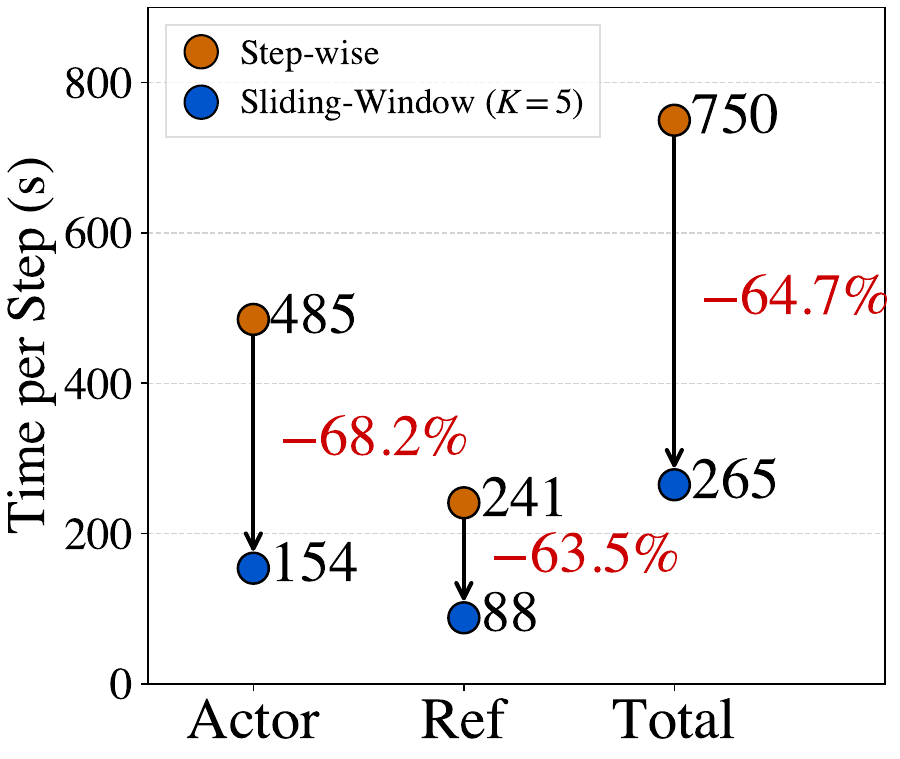}
  \end{minipage}
  \caption{(a) Frontier Sampling vs.\ uniform, DAPO, and curriculum learning: higher reward sustained by tracking the learning boundary. (b) Sliding-Window ($K{=}5$) vs.\ step-wise per-phase training step time; \textbf{2.83$\times$} end-to-end speedup.}
  \label{fig:rl_ablation}
\end{figure}

\noindent\textbf{Frontier Sampling.} We compare our frontier-based sampler against three baselines: uniform sampling, DAPO~\citep{yu2025dapo}, and curriculum learning. As shown in Figure~\ref{fig:rl_ablation}(a), uniform plateaus early on already-mastered or too-difficult tasks. DAPO improves via filtered rollouts, but its allocation does not adapt to per-task capability. Curriculum learning's fixed difficulty schedule fails to track the model's per-task success frontier. Frontier Sampling instead updates task priorities from EMA success rates and allocates more rollouts to tasks near the current learning boundary, sustaining a higher reward trajectory throughout training.

\begin{wrapfigure}{r}{0.4\textwidth}
  \centering
  \vspace{-\intextsep}
  \includegraphics[width=\linewidth]{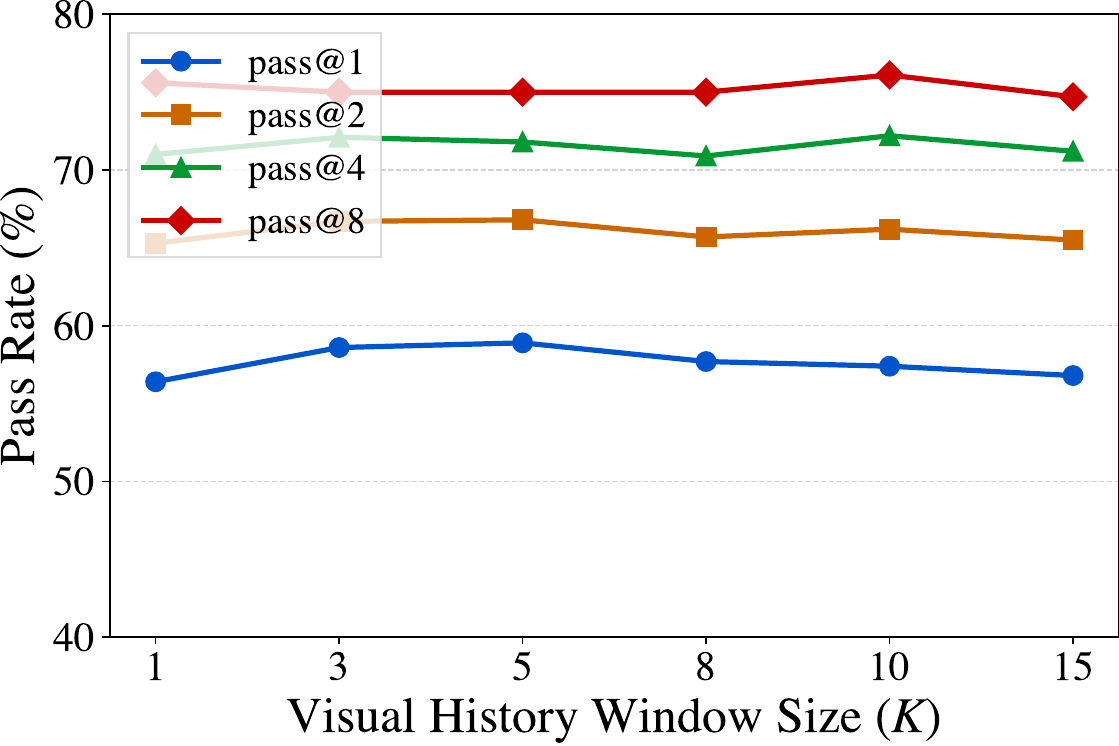}
  \caption{Pass@$k$ on OSWorld under different sliding-window sizes $K$ (Qwen3.5-9B). Bounded visual context does not reduce task success, and moderate windows ($K{=}3$--$5$) perform best.}
  \label{fig:sliding_passrate}
\end{wrapfigure}
\noindent\textbf{Visual Context Segmentation.} We compare Visual Context Segmentation ($K{=}5$) against the standard step-wise decomposition baseline on Qwen3.5-9B with batch size 32 and rollout-$n{=}8$, holding all other hyperparameters and the compute budget fixed. As shown in Figure~\ref{fig:rl_ablation}(b), it reduces the actor-update phase from 485s to 154s ($-$68.2\%) and the reference-policy phase from 241s to 88s ($-$63.5\%), bringing the per-step total from 750s to 265s---a \textbf{2.83$\times$} end-to-end speedup. The savings come from collapsing $O(nT)$ step-wise items into $O(nT/K)$ Sliding-Window segments and capping per-segment visual tokens, removing the long-tail rollout latency that dominates step-wise training. Crucially, this speedup does not come at the cost of quality: Figure~\ref{fig:sliding_passrate} reports pass@$k$ ($k{=}1,2,4,8$) on OSWorld across $K \in \{1,3,5,8,10,15\}$, where moderate windows ($K{=}3$--$5$) perform best---pass@1 peaks at 58.9\% ($K{=}5$) versus 56.4\% ($K{=}1$) and 56.8\% ($K{=}15$)---confirming that bounded visual context preserves task success while enabling the speedup.

\subsection{Analysis of \datapipelinename{}}
\label{sec:verisynth_analysis}

\begin{figure}[t]
\centering
\begin{minipage}[t]{0.54\linewidth}
  \centering
  \includegraphics[width=\linewidth]{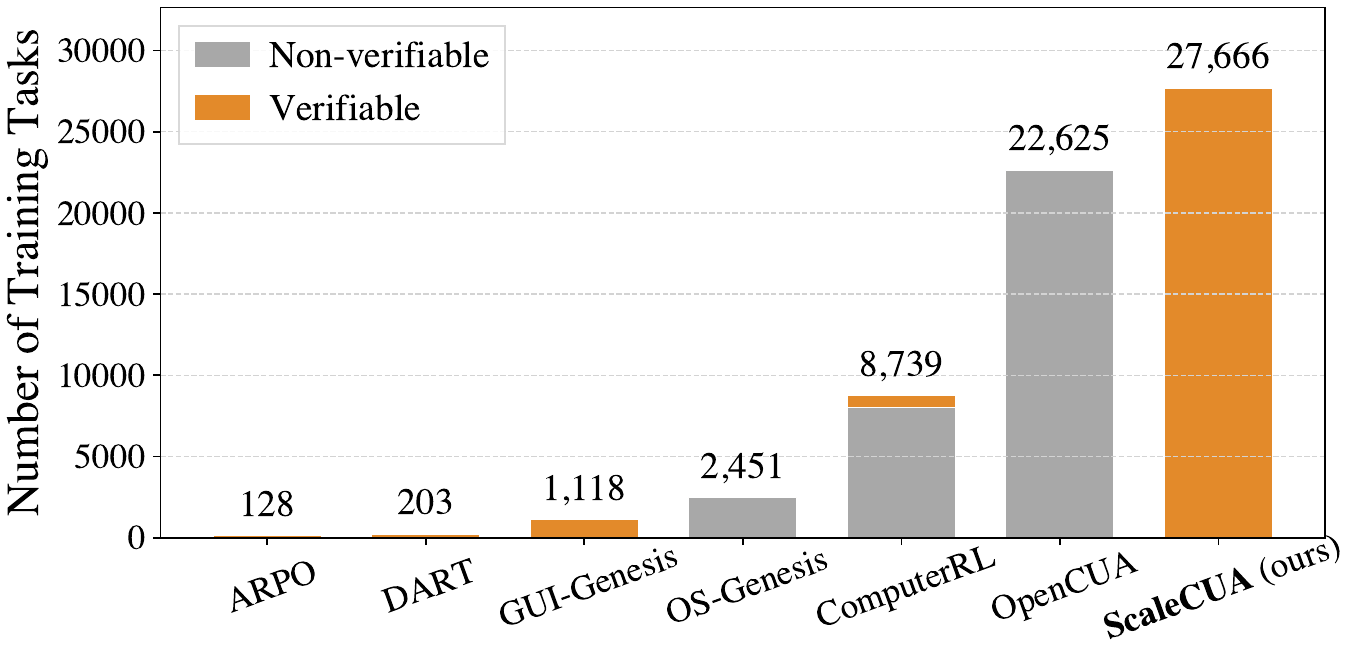}
\end{minipage}\hfill
\begin{minipage}[t]{0.42\linewidth}
  \centering
  \includegraphics[width=\linewidth]{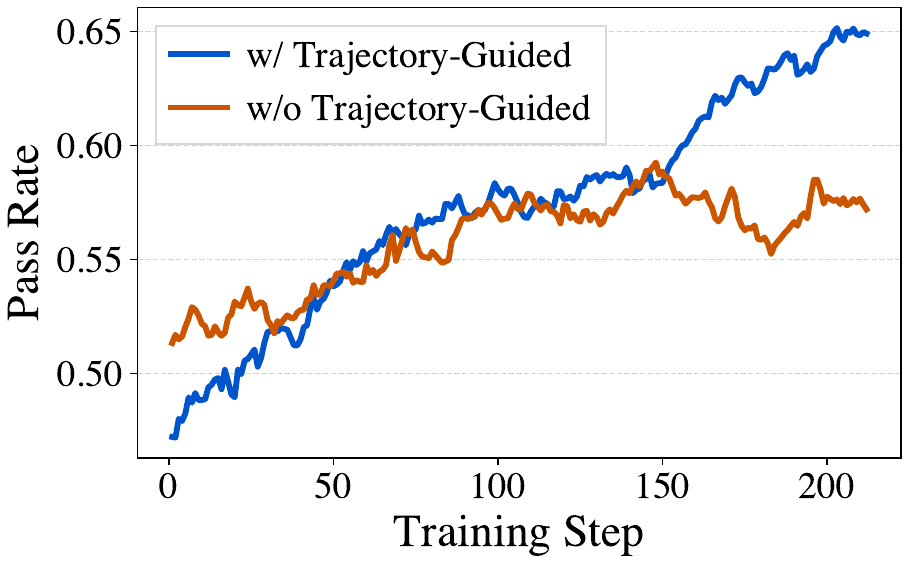}
\end{minipage}
\caption{\datapipelinename{} analysis. \textbf{Left:} \datapipelinename{} produces an order-of-magnitude larger verifiable task pool than prior desktop CUA methods. \textbf{Right:} augmenting the RL pool with trajectory-guided tasks consistently raises task-level reward throughout training.}
\label{fig:verigen_analysis}
\end{figure}

Running \datapipelinename{} at its 100+ parallel-worker scale on OSWorld produces over 24K candidate tasks, of which nearly 3K pass all filtering stages to form the final RL task pool.
This pool is more than an order of magnitude larger than what prior desktop CUA methods produce for either supervised warm-up or RL (Figure~\ref{fig:verigen_analysis}, left), placing \methodname{} on a distinctly different point of the data-scale axis.
The generated tasks span all 10 OSWorld application domains and all 6 ScienceBoard scientific domains (per-domain distribution and end-to-end synthesis cost in Appendix~\ref{app:autogen_filter_dist}).

We further evaluate the quality and training utility of the generated tasks from two angles.
For judge correctness, an expert human-judge audit on 10 trajectories per application domain shows that the executable judges agree with human labels on \textbf{82.5\%} of cases (details in Appendix~\ref{app:human_judge_audit}), indicating that the reward signals are not only mechanically executable but also broadly consistent with expert assessment.
For training utility, augmenting the RL pool with trajectory-guided tasks raises the OSWorld test score from 64.6\% to \textbf{68.7\%} (and the training-subset pass rate as shown in Figure~\ref{fig:verigen_analysis}, right), confirming the practical utility of \datapipelinename{} beyond raw task count.

\subsection{Case Study: Impact of Visual Context Window}
\label{sec:case_study}

To qualitatively illustrate why the window size $K$ matters beyond aggregate metrics, we examine a representative multi-app task: \emph{``Save all open Chrome blog articles as PDFs to \texttt{/home/user/Documents/Blog}, named by article title.''}
This task requires persistent awareness of the global goal (target directory, title-based naming) while executing a multi-step save workflow across browser tabs.
We evaluate the checkpoint (Qwen3.5-9B~\citep{qwen32025}) with $K \in \{1, 3, 5, 8, 15\}$ using 8 rollouts each.
A clear inverted-U pattern emerges: $K{=}5$ achieves 7/8 pass rate, $K{=}3$ reaches 5/8, while $K{=}1$, $K{=}8$, and $K{=}15$ all score 0/8.
With $K{=}1$, the agent completes the save workflow mechanically but fails to rename files in these rollouts---it loses track of the naming requirement after a few steps and saves with default Chrome filenames.
With $K{\geq}8$, stale visual context from early-episode actions (terminal commands, file manager dialogs) dilutes attention, causing the agent to enter repetitive action loops and inflating execution time by $2{\times}$--$3{\times}$.
$K{=}5$ retains exactly the right amount of recent history---enough to remember the target directory, naming convention, and remaining tabs, without interference from irrelevant earlier screenshots.
This confirms that the optimal $K$ identified by training speed analysis also maximizes task success: moderate visual context preserves essential state without overwhelming the model's attention capacity.

% Section 6: Conclusion
\section{Conclusion}
\label{sec:conclusion}

We presented \methodname{}, a unified framework that scales online RL for computer use agents through verifiable task synthesis and efficient online RL.
On the data side, \datapipelinename{} drives LLMs to interact with live OS containers through an iterative multi-agent loop that validates each judge by real execution. Trajectory-guided synthesis recycles rollout outcomes into new RL tasks, and a shared docker probe scales the pipeline to 100+ concurrent workers.
On the online RL side, we propose Frontier Sampling to direct rollouts to tasks at the model's current capability frontier via per-task EMA success rates, and propose Visual Context Segmentation to maintain a token-ID stream with a sliding visual window, balancing rollout and training engine pressure on long-horizon multimodal trajectories.
We hope \methodname{} offers a reusable recipe for verifiable-reward agent training and accelerates progress on open-source computer use agents.

\noindent\textbf{Limitations.} \emph{(i)} Each episode is capped at 50 interaction turns, which covers most OSWorld and ScienceBoard tasks but may underrepresent ultra-long workflows. \emph{(ii)} Our evaluation environments are Ubuntu-based desktops; generalization to Windows and macOS application stacks is left to future work. \emph{(iii)} We validate \methodname{} on 8B--9B-class vision-language models; behavior at substantially larger or smaller scales remains to be characterized.

\bibliographystyle{plainnat}

\bibliography{references}

%%%%%%%%%%%%%%%%%%%%%%%%%%%%%%%%%%%%%%%%%%%%%%%%%%%%%%%%%%%%

\appendix
\raggedbottom % keep pages from stretching glue; prevents huge appendix gaps

% Appendix
\section*{Appendix}

% --- Local layout tightening for the appendix only ---
\raggedbottom % stop LaTeX from stretching glue to fill pages
\setlength{\textfloatsep}{8pt plus 2pt minus 2pt}
\setlength{\intextsep}{8pt plus 2pt minus 2pt}
\setlength{\floatsep}{8pt plus 2pt minus 2pt}
\setlength{\abovecaptionskip}{4pt}
\setlength{\belowcaptionskip}{2pt}
\setlist{topsep=2pt, partopsep=0pt, parsep=2pt, itemsep=1pt}
\renewcommand{\arraystretch}{1.05}
\titlespacing*{\section}{0pt}{10pt plus 2pt minus 2pt}{4pt}
\titlespacing*{\subsection}{0pt}{6pt plus 1pt minus 1pt}{2pt}
\titlespacing*{\paragraph}{0pt}{4pt plus 1pt minus 1pt}{0.6em}

\section{Action Space and System Prompt}
\label{app:action_space}
Our agent receives screenshot observations and interacts through each benchmark's exposed \texttt{computer} action interface. We do not use accessibility trees or privileged environment state for perception. Below we provide the complete action space and system prompts for both evaluation environments.

\subsection{OSWorld Action Space}
The agent receives screenshots at each turn and issues actions through the following interface:
\begin{itemize}[leftmargin=*, itemsep=1pt, topsep=2pt, parsep=1pt]
    \item \textbf{Mouse actions:} \texttt{left\_click}, \texttt{right\_click}, \texttt{middle\_click}, \texttt{double\_click}, \texttt{triple\_click}, \texttt{left\_click\_drag}, \texttt{mouse\_move}, \texttt{left\_mouse\_down}, \texttt{left\_mouse\_up} --- each parameterized by \texttt{(x, y)} pixel coordinates.
    \item \textbf{Keyboard actions:} \texttt{key} (press a key or combination, e.g., \texttt{ctrl+s}), \texttt{hold\_key} (hold for a duration), \texttt{type} (type a string of text).
    \item \textbf{Scroll:} \texttt{scroll} with direction (\texttt{up/down/left/right}) and amount.
    \item \textbf{Utility:} \texttt{screenshot} (capture current screen), \texttt{wait} (pause for a duration), \texttt{cursor\_position} (get current cursor location).
\end{itemize}
Coordinates use a relative $1000 \times 1000$ grid that is rescaled to the actual screen resolution ($1280 \times 800$) by the environment harness.

\subsection{OSWorld System Prompt}
\begin{tcolorbox}[colback=gray!5!white, colframe=blue!75!black,
title=System Prompt for OSWorld, boxrule=0.3mm, arc=2mm,
auto outer arc=true, fontupper=\small,
left=4pt, right=4pt, top=3pt, bottom=3pt, boxsep=2pt,
before skip=4pt, after skip=4pt]
\texttt{<SYSTEM\_CAPABILITY>}\\[2pt]
You are utilising an Ubuntu virtual machine using x86\_64 architecture with internet access. You can install Ubuntu applications with your bash tool. Use curl instead of wget. To open browser, please just click on the Chrome icon. Using bash tool you can start GUI applications, but you need to set \texttt{export DISPLAY=:1} and use a subshell. GUI apps run with bash tool will appear within your desktop environment. When viewing a page, zoom out or scroll down to see everything before deciding something isn't available. DO NOT ask users for clarification during task execution. The screen's resolution is $1000 \times 1000$. Home directory is \texttt{/home/user}. Password for sudo is \texttt{password}.\\[2pt]
\texttt{</SYSTEM\_CAPABILITY>}
\end{tcolorbox}

\subsection{ScienceBoard Action Space}
The ScienceBoard environment extends the OSWorld action space with three additional actions for scientific workflow support:
\begin{itemize}[leftmargin=*, itemsep=1pt, topsep=2pt, parsep=1pt]
    \item \texttt{done}: Signal task completion (for action-oriented tasks).
    \item \texttt{answer}: Submit a computed value (for measurement/computation tasks), with the answer in the \texttt{text} field.
    \item \texttt{write\_file}: Write content to a file on the VM (file path in \texttt{text}, content in \texttt{content} field), supporting Unicode. This avoids the need to open a terminal for file operations.
\end{itemize}
All other mouse, keyboard, scroll, and utility actions are identical to OSWorld.

\subsection{ScienceBoard System Prompt}
\begin{tcolorbox}[colback=gray!5!white, colframe=blue!75!black,
title=System Prompt for ScienceBoard, boxrule=0.3mm, arc=2mm,
auto outer arc=true, fontupper=\small,
left=4pt, right=4pt, top=3pt, bottom=3pt, boxsep=2pt,
before skip=4pt, after skip=4pt]
\texttt{<SYSTEM\_CAPABILITY>}\\[2pt]
You are utilising an Ubuntu virtual machine with scientific software installed. The machine runs ScienceBoard applications: KAlgebra, Celestia, ChimeraX, GrassGIS, TeXstudio, and Lean/VS Code. You interact with the desktop via mouse and keyboard actions. You can see the screen through screenshots and must use GUI interaction to complete tasks.\\[2pt]
\texttt{</SYSTEM\_CAPABILITY>}\\[3pt]
\texttt{<APPLICATION\_HINTS>}
\begin{itemize}[leftmargin=*, itemsep=0pt, parsep=0.15em, topsep=0.15em]
    \item \textbf{KAlgebra}: mathematical calculator; use \texttt{:=} for variable assignment.
    \item \textbf{Celestia}: space simulator; use GUI menus.
    \item \textbf{ChimeraX}: molecular visualization; use command line at bottom.
    \item \textbf{GrassGIS}: GIS tool; use GUI and terminal for GRASS commands.
    \item \textbf{TeXstudio}: \LaTeX{} editor; use Ctrl+R for Find \& Replace.
    \item \textbf{Lean/VS Code}: theorem prover in Lean 4 with Mathlib; write proofs via \texttt{write\_file} action.
\end{itemize}
\texttt{</APPLICATION\_HINTS>}
\end{tcolorbox}

\section{\datapipelinename{} Pipeline Details}
\label{app:autogen_details}
This section provides the full set of \datapipelinename{} details referenced from the main text, including an anonymized example of a generated task, the multi-agent filtering ablation, the per-domain task distribution, judge-correctness audits, capability-ordered validation on held-out tasks, the utility of trajectory-guided augmentation, and an overlap audit against benchmark evaluation pools.

\subsection{Example Generated Task Schema}
\label{app:verigen_task_example}
Table~\ref{tab:verigen_task_example} shows an anonymized example of a \datapipelinename{}-generated task. The concrete task content, identifiers, URLs, and local paths are masked, while the schema fields are preserved to illustrate how each task couples an executable setup with a task-specific Python judge.

\begin{table}[!htbp]
\centering
\caption{An anonymized \datapipelinename{} task example. The example preserves the generated schema while masking task-specific identifiers, file names, URLs, and local paths.}
\label{tab:verigen_task_example}
\small
\setlength{\tabcolsep}{5pt}
\renewcommand{\arraystretch}{1.08}
\begin{tabular}{p{0.20\linewidth}p{0.74\linewidth}}
\toprule
\textbf{Field} & \textbf{Anonymized Content} \\
\midrule
\texttt{id} & \texttt{<task-uuid>} \\
\texttt{new\_id} & \texttt{<task-uuid>\_<augmentation-tag>} \\
\texttt{snapshot} & \texttt{<desktop-application-snapshot>} \\
\texttt{instruction} & Apply a specified presentation-wide style change to all textual content in a desktop office document. \\
\texttt{source} & \texttt{<public-help-page-or-application-reference>} \\
\texttt{setup config} & Download \texttt{<input-file>} to the VM desktop and open it with the target application. \\
\texttt{related\_apps} & \texttt{[<target-desktop-application>]} \\
\midrule
\texttt{postconfig} & Activate the target application window, wait for UI stabilization, execute a save action, and wait until the file state is persisted before judging. \\
\texttt{judge func} & \texttt{check\_<app>\_<property>\_\_<hash>} \\
\texttt{result getter} & Inspect the final environment state, such as document metadata, application state, accessibility-tree properties, or generated files. \\
\texttt{expected rule} & A task-specific rule, e.g., the final document must satisfy a specified formatting or state constraint. \\
\texttt{reward} & A deterministic scalar score returned by the Python judge, typically binary for all-or-nothing tasks and dense when sub-goals are available. \\
\midrule
\texttt{metadata} & Natural-language explanation, generated getter/metric paths with masked local roots, estimated difficulty, estimated steps, augmentation type, and validation records. \\
\texttt{validation} & LLM-judge pass flag and score, rule-validation pass flag, execution-in-the-loop pass flag, and optional repair suggestions. \\
\bottomrule
\end{tabular}
\end{table}

Figure~\ref{fig:verigen_judge_example} shows the corresponding anonymized Python judge. The getter retrieves the final artifact from the environment, extracts task-specific state variables, and the metric converts the extracted state into a deterministic reward.

\begin{figure}[!htbp]
\centering
\begin{tcolorbox}[colback=gray!5!white, colframe=blue!75!black,
title=\datapipelinename{} Judge Function (anonymized), boxrule=0.3mm, arc=2mm,
auto outer arc=true, left=4pt, right=4pt, top=2pt, bottom=2pt, boxsep=2pt,
before skip=4pt, after skip=4pt]
\begin{minipage}{\linewidth}
\footnotesize
\begin{verbatim}
def get_document_property__<hash>(env, config):
    """Extract task-specific properties from the final desktop artifact."""
    file_bytes = env.controller.get_file(config["path"])
    if not file_bytes:
        return set()
    with tempfile.NamedTemporaryFile(suffix="<artifact-ext>",
                                     delete=False) as tmp:
        tmp.write(file_bytes)
        tmp_path = tmp.name
    try:
        document = load_desktop_artifact(tmp_path)
        observed_values = set()
        for page in document.pages:
            for element in page.elements:
                if has_relevant_content(element):
                    value = extract_target_property(element)
                    if value:
                        observed_values.add(value)
        return observed_values
    finally:
        os.unlink(tmp_path)

def check_document_property__<hash>(result, expected, **options):
    """Return a deterministic reward for the task-specific rule."""
    expected_value = expected.get("<target_property>")
    if not result:
        return 0.0
    return 1.0 if result == {expected_value} else 0.0
\end{verbatim}
\end{minipage}
\end{tcolorbox}
\caption{Anonymized Python judge corresponding to the task schema in Table~\ref{tab:verigen_task_example}. The concrete artifact type, function suffix, target property, and file paths are masked, while the generated evaluator structure is preserved.}
\label{fig:verigen_judge_example}
\end{figure}

\subsection{Multi-Agent Filtering Ablation and Task Distribution}
\label{app:autogen_filter_dist}
Table~\ref{tab:autogen_role_ablation} (in the main text) ablates the \datapipelinename{} multi-agent filtering loop by removing each module. The full pipeline achieves 94.5\% executable rate for generated judge functions; removing the LLM Judge Agent causes the largest drop ($-$32.2\%), followed by the Fix Agent ($-$16.4\%) and Rule Validator ($-$8.3\%).

\begin{figure}[!htbp]
  \centering
  \includegraphics[width=0.95\linewidth]{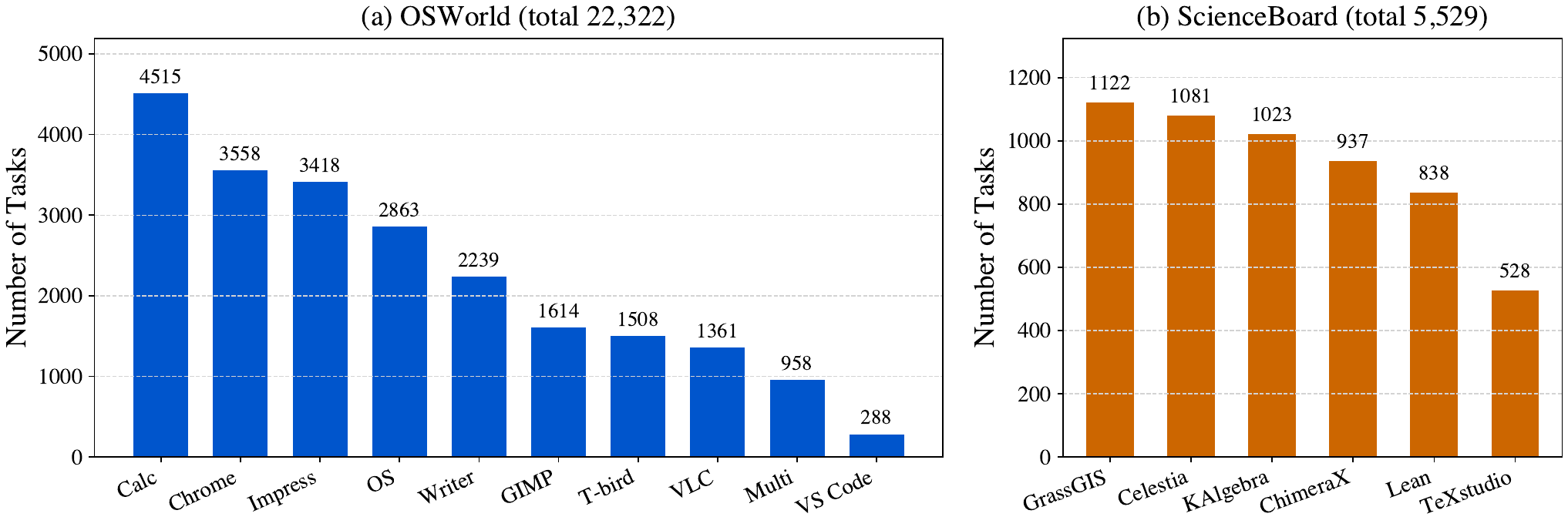}
  \caption{Per-domain task distribution of \datapipelinename{}-generated tasks. (a) OSWorld: 10 application domains, total 22{,}322 tasks. (b) ScienceBoard: 6 scientific domains, total 5{,}529 tasks. Counts aggregate across all generation stages.}
  \label{fig:autogen_per_domain}
\end{figure}

Figure~\ref{fig:autogen_per_domain} shows the per-domain distribution of \datapipelinename{}-generated tasks for OSWorld (10 application domains) and ScienceBoard (6 scientific domains), aggregating across all generation stages. Coverage spans every domain, with the largest volumes in office productivity (Calc, Impress, Writer) and browsing (Chrome) on OSWorld and balanced coverage across scientific applications on ScienceBoard.

\noindent\textbf{End-to-end synthesis cost.} Each synthesis-worker call generates three candidates in parallel and passes through rule validation, LLM-judge validation, and executable Docker validation. Across all runs, the measured executable rate is 94.5\%, with an average end-to-end latency of 221.7 seconds and 0.93--1.01 USD per accepted task. At 100+ parallel workers, this translates into a daily throughput of roughly 35K candidate tasks under steady-state operation.

\subsection{Expert Human-Judge Audit}
\label{app:human_judge_audit}
To directly characterize judge correctness, we randomly sample 10 \datapipelinename{}-generated trajectories from each application domain in OSWorld and ScienceBoard and ask expert human reviewers to inspect the trajectory evidence and final state. As shown in Table~\ref{tab:human_judge_audit}, executable judge decisions agree with expert human labels on 82.5\% of the 160 audited examples, with 78.0\% agreement on OSWorld and 90.0\% on ScienceBoard. This audit complements the executable-rate and model-capability validations by providing a direct estimate of judge-human agreement and residual false positive/false negative cases.

\begin{table}[!htbp]
\centering
\caption{Expert human-judge audit of \datapipelinename{}-generated trajectories. We randomly sample 10 examples per application domain from OSWorld and ScienceBoard. Human Valid reports the fraction judged successful by expert human reviewers. Agreement compares the executable judge decision with the human label; FP/FN count judge false positives and false negatives against the human label.}
\label{tab:human_judge_audit}
\resizebox{\textwidth}{!}{%
\begin{tabular}{l l r c c r r}
\toprule
\textbf{Benchmark} & \textbf{Domain} & \textbf{N} & \textbf{Human Valid (\%)} & \textbf{Agreement (\%)} & \textbf{FP} & \textbf{FN} \\
\midrule
OSWorld & Chrome & 10 & 70.0 & 70.0 & 1 & 2 \\
OSWorld & GIMP & 10 & 90.0 & 80.0 & 0 & 2 \\
OSWorld & LibreOffice Calc & 10 & 90.0 & 90.0 & 1 & 0 \\
OSWorld & LibreOffice Impress & 10 & 80.0 & 80.0 & 1 & 1 \\
OSWorld & LibreOffice Writer & 10 & 80.0 & 70.0 & 2 & 1 \\
OSWorld & Multi-Apps & 10 & 70.0 & 90.0 & 0 & 1 \\
OSWorld & OS & 10 & 100.0 & 60.0 & 0 & 4 \\
OSWorld & Thunderbird & 10 & 80.0 & 70.0 & 2 & 1 \\
OSWorld & VLC & 10 & 70.0 & 80.0 & 1 & 1 \\
OSWorld & VS Code & 10 & 90.0 & 90.0 & 1 & 0 \\
\midrule
\textbf{OSWorld} & \textbf{Overall} & \textbf{100} & \textbf{82.0} & \textbf{78.0} & \textbf{9} & \textbf{13} \\
\midrule
ScienceBoard & Celestia & 10 & 30.0 & 90.0 & 1 & 0 \\
ScienceBoard & ChimeraX & 10 & 80.0 & 100.0 & 0 & 0 \\
ScienceBoard & GrassGIS & 10 & 40.0 & 100.0 & 0 & 0 \\
ScienceBoard & KAlgebra & 10 & 100.0 & 80.0 & 0 & 2 \\
ScienceBoard & Lean & 10 & 50.0 & 80.0 & 0 & 2 \\
ScienceBoard & TeXstudio & 10 & 50.0 & 90.0 & 1 & 0 \\
\midrule
\textbf{ScienceBoard} & \textbf{Overall} & \textbf{60} & \textbf{58.3} & \textbf{90.0} & \textbf{2} & \textbf{4} \\
\midrule
\textbf{All} & \textbf{Overall} & \textbf{160} & \textbf{73.1} & \textbf{82.5} & \textbf{11} & \textbf{17} \\
\bottomrule
\end{tabular}%
}
\end{table}

\subsection{Capability-Ordered Task Validation}
\label{app:task_validation}
A complementary concern is whether the generated tasks are logically solvable, meaningful, and appropriately difficult for online RL. We evaluate two models of differing capability---Qwen3-VL-8B-Base and Claude-4.5-Sonnet---on a held-out subset of 150 tasks per augmentation type (task-only and trajectory-guided). As shown in Table~\ref{tab:task_validation}, Claude-4.5-Sonnet achieves a pass rate above 50\%, showing that many \datapipelinename{} tasks are solvable and that judge errors are not the dominant failure mode. The Base model achieves around 13\% pass rate, suggesting that tasks are non-trivial for weaker models while remaining solvable by stronger agents.

\begin{table}[!htbp]
\centering
\small
\caption{Validation of generated tasks across different model capabilities (150 tasks per type).}
\label{tab:task_validation}
\begin{tabular}{lcccc}
\toprule
\multirow{2}{*}{Model} & \multicolumn{2}{c}{Task Augmentation} & \multicolumn{2}{c}{Trajectory Augmentation} \\
\cmidrule(lr){2-3} \cmidrule(lr){4-5}
& Avg. Score & Pass Rate (\%) & Avg. Score & Pass Rate (\%) \\
\midrule
Qwen3-VL-8B-Base   & 13.0 & 12.6 & 19.0 & 14.0 \\
Claude-4.5-Sonnet  & 55.5 & 52.6 & 59.0 & 50.0 \\
\midrule
Gap (Sonnet$-$Base) & \multicolumn{2}{c}{\textbf{+40.0}} & \multicolumn{2}{c}{\textbf{+36.0}} \\
\bottomrule
\end{tabular}
\end{table}

\subsection{Trajectory-Guided Task Augmentation}
\label{app:traj_augmentation}
Figure~\ref{fig:verigen_analysis} (right, in the main text) compares RL training with and without trajectory-guided task augmentation, evaluated on the same capability-driven validation subset. Adding trajectory-guided tasks---derived from rollout experience as simpler variants of failed tasks or harder extensions of solved ones---consistently improves task-level reward throughout training, complementing the overall benchmark gain from 64.6\% to 68.7\% reported in the main text.

\subsection{Training vs. Evaluation Overlap Audit}
\label{app:overlap_audit}
To check whether \datapipelinename{} training tasks duplicate benchmark evaluation tasks, we audit the generated subsets used for training against the corresponding benchmark task pools. The audit separates provenance metadata from task-content overlap: source IDs indicate which seed environment or task inspired generation, but they are not treated as leakage unless the generated task also copies the objective or judge. We report exact full-JSON duplication, normalized-instruction duplication, objective near-duplicates using a composite instruction-similarity threshold of 0.90, and exact judge reuse.

As shown in Table~\ref{tab:overlap_audit}, the training subsets contain no full-JSON duplicates and no exact instruction duplicates. Objective near-duplicates are rare (0.19\% on OSWorld and 0.36\% on ScienceBoard). ScienceBoard has no exact reuse of evaluation judges; OSWorld shows a small amount of exact evaluator reuse (2.10\%), corresponding to reusable generic evaluator templates rather than copied task instructions. The ScienceBoard instruction-rewrite control subset intentionally reuses source evaluators and is excluded from training.

\begin{table}[!htbp]
\caption{Training-data overlap audit against benchmark task pools. We report only the generated subsets used for training. The ScienceBoard instruction-rewrite control subset, which intentionally reuses source evaluators, is excluded from training and omitted here.}
\label{tab:overlap_audit}
\centering
\resizebox{\textwidth}{!}{%
\begin{tabular}{l l r c c c c c}
\toprule
\textbf{Benchmark} & \textbf{Generated subset} & \textbf{N} & \textbf{Full JSON dup.} & \textbf{Instruction exact dup.} & \textbf{Objective near-dup.} & \textbf{Judge exact reuse} & \textbf{Objective sim. p50/p95} \\
\midrule
OSWorld & traj\_verify & 1049 & 0.00\% (0) & 0.00\% (0) & 0.19\% (2) & 2.10\% (22) & 0.319 / 0.628 \\
ScienceBoard & sci\_traj\_verify & 825 & 0.00\% (0) & 0.00\% (0) & 0.36\% (3) & 0.00\% (0) & 0.342 / 0.596 \\
\bottomrule
\end{tabular}%
}
\end{table}

\section{Sliding-Window Details}
\label{app:sliding_window}
This section provides implementation details for Sliding-Window Trajectory Segmentation: the token-stream update algorithm, the ablation configurations used in the main-text compute sweep, and a finer-grained view of how $K$ affects task success.

\subsection{Trajectory Processing Algorithm}
\label{app:sliding_algorithm}
\paragraph{Token-ID stream.}
Policy-gradient training requires the rollout log-probabilities and training tokens to stay aligned, which can be broken if the same history is re-tokenized under different BPE boundaries. We therefore maintain the stream incrementally with \emph{token diffs}: response tokens and their log-probabilities are appended as trainable tokens, while new observation tokens are appended with zero loss mask. After pruning, the retained history is re-tokenized with a zero loss mask, so previous assistant responses only provide context in later segments and their losses are not recomputed. Algorithm~\ref{alg:sliding_window} gives the full procedure; all segments from the same rollout share the trajectory-level reward and advantage.

\begin{algorithm}[!htbp]
\caption{Sliding-window trajectory processing with token-ID stream}
\label{alg:sliding_window}
\begin{algorithmic}[1]
\Require Episode with $T$ turns; $K$ = images to keep; $\Delta$ = min removal threshold
\Ensure Set of trace segments $\mathcal{S}$
\State $\mathcal{S} \leftarrow \emptyset$; \; $n_{\text{img}} \leftarrow 0$
\State Tokenize initial history $\mathcal{H}$ $\rightarrow$ token-ID stream $(\mathbf{ids}, \mathbf{mask}, \mathbf{logp})$ with $\mathbf{mask}{=}\mathbf{0}$
\For{$t = 1, \ldots, T$}
    \State Generate response tokens $(\mathbf{ids}^{\text{gen}}_t, \mathbf{logp}^{\text{gen}}_t)$ from policy
    \State $\mathbf{ids} \mathrel{+}= \mathbf{ids}^{\text{gen}}_t$; \; $\mathbf{mask} \mathrel{+}= \mathbf{1}$; \; $\mathbf{logp} \mathrel{+}= \mathbf{logp}^{\text{gen}}_t$ \Comment{Trainable tokens}
    \State Execute action, receive observation with screenshot; \; $n_{\text{img}} \leftarrow n_{\text{img}} + 1$
    \If{$n_{\text{img}} > K + \Delta$} \Comment{Pruning triggered}
        \State Save segment: $\mathcal{S} \leftarrow \mathcal{S} \cup \{(\mathbf{ids}, \mathbf{mask}, \mathbf{logp})\}$
        \State Remove oldest $\Delta$ images from $\mathcal{H}$; \; $n_{\text{img}} \leftarrow n_{\text{img}} - \Delta$
        \State Re-tokenize $\mathcal{H}$ $\rightarrow$ $(\mathbf{ids}, \mathbf{mask}, \mathbf{logp}) \leftarrow (\text{new\_ids}, \mathbf{0}, \mathbf{0})$ \Comment{Fresh stream}
    \Else \Comment{Token-diff append}
        \State $l \leftarrow |\textsc{Tokenize}(\mathcal{H})|$ \Comment{Length before appending $o_t$}
        \State $\mathbf{diff} \leftarrow \textsc{Tokenize}(\mathcal{H} \oplus o_t)[l:]$ \Comment{New obs tokens only}
        \State $\mathbf{ids} \mathrel{+}= \mathbf{diff}$; \; $\mathbf{mask} \mathrel{+}= \mathbf{0}$; \; $\mathbf{logp} \mathrel{+}= \mathbf{0}$ \Comment{Non-trainable tokens}
    \EndIf
\EndFor
\State Save final stream: $\mathcal{S} \leftarrow \mathcal{S} \cup \{(\mathbf{ids}, \mathbf{mask}, \mathbf{logp})\}$
\State Assign trajectory reward $r$ to all segments in $\mathcal{S}$
\State \Return $\mathcal{S}$
\end{algorithmic}
\end{algorithm}

\subsection{Training Ablation Configurations}
\label{app:sliding_configs}
Table~\ref{tab:sliding_configs} reports the rollout-to-training worker ratio used for each combination of sliding-window size $K$ and compute scale in the ablation study (Section~\ref{sec:ablation}). For all configurations, the batch size is fixed at 16 distinct trajectory UIDs and rollout-$n$ at 8. Smaller $K$ values produce more segments per episode with shorter sequences, so fewer rollout workers and more training workers are allocated. Larger $K$ values produce fewer but longer segments, requiring more rollout workers to keep the training engine fed. In practice, 4-node and 8-node setups can use larger batch sizes, which would further amplify the U-shaped trend in Figure~\ref{fig:rl_ablation}(b) by increasing training-engine pressure at small $K$ and rollout-engine pressure at large $K$; the relatively flat 8-node curve at large $K$ reflects that batch size 16 underutilizes the available parallelism and partially masks the rollout-engine bottleneck.

\begin{table}[!htbp]
\caption{Rollout-to-training node allocation for sliding-window training ablation experiments. All runs use Qwen3.5-9B~\citep{qwen32025} with batch size 16 and rollout-$n$=8. Rollout ratio denotes the fraction of total nodes dedicated to rollout.}
\label{tab:sliding_configs}
\centering
\begin{tabular}{l c c c c c}
\toprule
\textbf{$K$} & \textbf{Total Nodes} & \textbf{Rollout Ratio} & \textbf{Rollout Nodes} & \textbf{Training Nodes} & \textbf{Avg Step Time (s)} \\
\midrule
\multicolumn{6}{l}{\emph{2 Nodes (16 GPUs)}} \\
1  & 2 & 0.25 & 0.5 & 1.5 & 1102.6 \\
3  & 2 & 0.25 & 0.5 & 1.5 & 643.8 \\
5  & 2 & 0.50 & 1.0 & 1.0 & 703.4 \\
8  & 2 & 0.50 & 1.0 & 1.0 & 687.5 \\
10 & 2 & 0.50 & 1.0 & 1.0 & 1042.6 \\
15 & 2 & 0.50 & 1.0 & 1.0 & 1649.3 \\
\midrule
\multicolumn{6}{l}{\emph{4 Nodes (32 GPUs)}} \\
1  & 4 & 0.25 & 1 & 3 & 962.0 \\
3  & 4 & 0.25 & 1 & 3 & 525.8 \\
5  & 4 & 0.25 & 1 & 3 & 405.2 \\
8  & 4 & 0.50 & 2 & 2 & 400.2 \\
10 & 4 & 0.75 & 3 & 1 & 612.6 \\
15 & 4 & 0.75 & 3 & 1 & 693.3 \\
\midrule
\multicolumn{6}{l}{\emph{8 Nodes (64 GPUs)}} \\
1  & 8 & 0.25 & 2 & 6 & 502.9 \\
3  & 8 & 0.25 & 2 & 6 & 279.4 \\
5  & 8 & 0.50 & 4 & 4 & 302.9 \\
8  & 8 & 0.50 & 4 & 4 & 282.5 \\
10 & 8 & 0.50 & 4 & 4 & 434.8 \\
15 & 8 & 0.75 & 6 & 2 & 519.9 \\
\bottomrule
\end{tabular}
\end{table}

\section{Training Details}
\label{app:training_details}
Table~\ref{tab:rl_hparams} summarizes the hyperparameters used in the RL stage.

\noindent\textbf{RL stage.} Online RL is conducted using AgentRL~\citep{agentrl2025} with GRPO~\citep{shao2024deepseekmath} as the base algorithm. We train for up to 1,000 iterations with a learning rate of $3 \times 10^{-6}$ (no weight decay). Each iteration collects $n{=}8$ rollouts per task with a batch size of 32 tasks (16 for ablation studies) and a concurrency of 600 parallel VM environments. The ViT backbone is frozen during RL; only the language model parameters are updated. The rollout engine uses vLLM with TP=2, while the training engine uses Megatron with TP=4. Rollout sampling temperature is 0.8. Each episode runs for a maximum of 50 turns.

We adopt asymmetric clipping from DAPO~\citep{yu2025dapo}: $\epsilon_l{=}0.2$, $\epsilon_h{=}0.28$, with a clip ratio cap of $c{=}3.0$. A KL penalty $\lambda{=}1 \times 10^{-4}$ regularizes against the reference policy. Incomplete episodes (agent fails to finish within the turn limit) receive a reward penalty of $-0.2$. For Frontier Sampling, tasks are ranked by Eq.~\eqref{eq:sampling_weight}; the top $3{\times}B$ tasks form the frontier candidate pool, $(1{-}\gamma)B$ tasks are sampled from this pool according to $w_i$, and $\gamma B$ tasks are sampled uniformly from outside the pool.

\noindent\textbf{RL objective.} At each iteration, Frontier Sampling selects a task batch $\mathcal{B}$ according to Eq.~\eqref{eq:sampling_weight}. For each task $i \in \mathcal{B}$, the rollout policy $\pi_{\text{rollout}}$ generates $n$ trajectories, yielding rewards $\{r_{i,j}\}$ from judge functions and trace segment sets $\{\mathcal{S}_{i,j}\}$ from Visual Context Segmentation. The group-relative advantage $\hat{A}_{i,j}$ is computed over the $n$ rollouts of each task and shared across all segments of the same rollout, while loss masks ensure each assistant response is optimized exactly once:
\begin{equation}
    \mathcal{L}(\theta) = -\mathbb{E}_{i \sim \mathcal{B}} \; \mathbb{E}_{j=1}^{n} \; \sum_{s \in \mathcal{S}_{i,j}} \sum_{t \in s} \left[ \min\!\Big( \rho_t \, \hat{A}_{i,j},\; \text{clip}(\rho_t, 1{-}\epsilon_l, 1{+}\epsilon_h) \, \hat{A}_{i,j} \Big) \right] + \lambda \, D_{\text{KL}}\!\left(\pi_\theta \,\|\, \pi_{\text{ref}}\right),
    \label{eq:loss}
\end{equation}
where $\rho_t = \pi_\theta(a_t \mid x_{<t}) / \pi_{\text{rollout}}(a_t \mid x_{<t})$ is the importance sampling ratio, $\epsilon_l$ and $\epsilon_h$ are the lower and upper clipping ranges, $\lambda$ is the KL coefficient, and the inner sum runs over trainable tokens ($\mathbf{mask}_t{=}1$) within each segment.

\begin{table}[!htbp]
\centering
\small
\caption{RL hyperparameters.}
\label{tab:rl_hparams}
\setlength{\tabcolsep}{6pt}
\begin{tabular}{ll}
\toprule
\textbf{Hyperparameter} & \textbf{Value} \\
\midrule
Algorithm & GRPO \\
Learning rate & $3 \times 10^{-6}$ \\
Clip ratio (low / high) & 0.2 / 0.28 \\
Clip ratio cap $c$ & 3.0 \\
KL penalty $\lambda$ & $1 \times 10^{-4}$ \\
Entropy coefficient & 0.0 \\
Batch size (tasks per iter) & 32 \\
Rollouts per task $n$ & 8 \\
Max iterations & 1,000 \\
Rollout temperature & 0.8 \\
Max turns per episode & 50 \\
Max sequence length & 40,000 \\
Incomplete episode penalty & $-0.2$ \\
\midrule
\multicolumn{2}{l}{\emph{Sampling hyperparameters}} \\
Target success rate $\mu$ & 0.5 \\
Bandwidth $\sigma$ & 0.25 \\
Random sample ratio $\gamma$ & 0.2 \\
EMA smoothing $\alpha$ & 0.2 \\
Initial pass rate range $[p_{\min}, p_{\max}]$ & $[0.0, 1.0]$ \\
\midrule
\multicolumn{2}{l}{\emph{Sliding-window training hyperparameters}} \\
Images to keep $K$ (OSWorld) & 5 \\
Images to keep $K$ (ScienceBoard) & 8 \\
Min removal threshold $\Delta$ & 5 \\
\bottomrule
\end{tabular}
\end{table}

\noindent\textbf{Infrastructure and compute.} The main RL experiments use 8 nodes, each equipped with 8 NVIDIA H800 (80\,GB) GPUs (64 GPUs total). The rollout engine is allocated 25\% of the GPU resources (16 GPUs) and the remaining GPUs are used by the actor/reference training engine. Table~\ref{tab:training_compute} reports the training- and environment-side compute profiles in a single panel. Checkpoints are saved every 25 iterations.

\begin{table}[!htbp]
\centering
\small
\setlength{\tabcolsep}{4pt}
\renewcommand{\arraystretch}{1.15}
\caption{Compute profile for the RL stage. \textbf{Left:} training-side configuration; average step time is computed from the logged \texttt{timings/step} metric of representative production runs. \textbf{Right:} environment-side rollout configuration; environment workers are Docker containers, matched one-to-one with rollout workers.}
\label{tab:training_compute}
\begin{minipage}[t]{0.52\linewidth}
\centering
\begin{tabular}{lcc}
\toprule
\textbf{Item} & \textbf{OSWorld RL} & \textbf{ScienceBoard RL} \\
\midrule
GPU cluster & \multicolumn{2}{c}{8 H800 nodes, 64 GPUs} \\
Rollout ratio & \multicolumn{2}{c}{0.25} \\
Rollout GPUs & \multicolumn{2}{c}{16 GPUs} \\
Actor/ref GPUs & \multicolumn{2}{c}{48 GPUs} \\
Rollout engine & \multicolumn{2}{c}{vLLM, TP=2} \\
Training engine & \multicolumn{2}{c}{Megatron-LM, TP=4} \\
Task batch size & 32 & 16 \\
Rollouts per task & 8 & 8 \\
Env.\ concurrency & 600 & 300 \\
Sampling $\sigma$ & 0.25 & 0.15 \\
Visual window $K$ & 5 & 8 \\
Avg.\ step time & 672.1\,s & 379.0\,s \\
\bottomrule
\end{tabular}
\end{minipage}%
\hfill
\begin{minipage}[t]{0.46\linewidth}
\centering
\begin{tabular}{ll}
\toprule
\textbf{Item} & \textbf{Value} \\
\midrule
Env.\ backend & OSWorld / SciBoard \\
Peak concurrency & 600 containers \\
Worker mapping & 1 container / worker \\
Avg.\ rollout time & 539.9\,s \\
Env.\ lifecycle & deploy / interact / clean \\
Action horizon & $\le$ 50 turns / rollout \\
\bottomrule
\end{tabular}
\end{minipage}
\end{table}

%%%%%%%%%%%%%%%%%%%%%%%%%%%%%%%%%%%%%%%%%%%%%%%%%%%%%%%%%%%%

\end{document}